\documentclass[%
 aip,cha,
 amsmath,amssymb,
 reprint,%
]{revtex4-2}

\usepackage{graphicx}
\usepackage{dcolumn}
\usepackage{bm}

\usepackage[utf8]{inputenc}
\usepackage[T1]{fontenc}
\usepackage{mathptmx}
\usepackage{etoolbox}

\usepackage{comment}
\usepackage{algorithm}
\usepackage{algpseudocode}
\usepackage[margin=1in]{geometry}
\usepackage{graphicx}
\usepackage{amssymb}
\usepackage{amsmath}
\usepackage{mathrsfs}
\usepackage{placeins}
\usepackage{caption3}
\usepackage{wrapfig}
\usepackage{caption} 
\usepackage{subcaption}
\usepackage{xcolor}
\usepackage{contour}
\usepackage{color}
\usepackage{outlines}
\usepackage[bookmarksnumbered=true,breaklinks=true,colorlinks,citecolor=blue,linkcolor=blue]{hyperref}
\usepackage{soul} 
\usepackage[font=small,labelfont=bf,
justification=raggedright,
format=plain]{caption} 
\setstcolor{red}

\usepackage[normalem]{ulem}
\newcommand{\MDG}[1]{\textcolor{magenta}{*** #1 ***}}
\renewcommand{\MDG}[1]{}
\newcommand{\MDGrevise}[1]{\textcolor{black}{#1}}
\definecolor{ao(english)}{rgb}{0.0, 0.5, 0.0}
\newcommand{\AG}[1]{\textcolor{black}{#1}}
\newcommand{\AGrevise}[1]{\textcolor{black}{#1}}

\newcommand{\Enc}{\mathcal{E}}
\newcommand{\Dec}{\mathcal{D}}

\newcommand{\FOM}{\mathrm{FOM}}
\newcommand{\MGN}{\text{MG-N}}

\usepackage{xcolor}

\usepackage{array}
\newcolumntype{W}{>{\centering\arraybackslash}p{0.28\linewidth}}
\newcolumntype{N}{>{\centering\arraybackslash}p{0.15\linewidth}}

\newcolumntype{a}{>{\centering\arraybackslash}p{0.14\linewidth}}
\newcolumntype{j}{>{\centering\arraybackslash}p{0.18\linewidth}}
\newcolumntype{z}{>{\centering\arraybackslash}p{0.41\linewidth}}
\newcolumntype{d}{>{\centering\arraybackslash}p{0.21\linewidth}}

\newcolumntype{e}{>{\centering\arraybackslash}p{0.15\linewidth}}
\newcolumntype{f}{>{\centering\arraybackslash}p{0.22\linewidth}}
\newcolumntype{g}{>{\centering\arraybackslash}p{0.31\linewidth}}
\newcolumntype{h}{>{\centering\arraybackslash}p{0.24\linewidth}}

\usepackage{setspace}

\newcommand{\specialcell}[2][c]{\begin{tabular}[#1]{@{}c@{}}#2\end{tabular}}

\begin{document}

\title{Blending data and physics for reduced-order modeling of systems with spatiotemporal chaotic dynamics}



\author{Alex Guo}
\affiliation{Department of Chemical and Biological Engineering, University of Wisconsin-Madison, Madison WI 53706, USA}

\author{Michael D. Graham}\email{mdgraham@wisc.edu}
\affiliation{Department of Chemical and Biological Engineering, University of Wisconsin-Madison, Madison WI 53706, USA}


\date{\today}

\begin{abstract} 
While data-driven techniques are powerful tools for reduced-order modeling of systems with chaotic dynamics, great potential remains for leveraging known physics (i.e.~a full-order model (FOM)) to improve predictive capability.
We develop a hybrid reduced order model (ROM), informed by both data and FOM, for evolving spatiotemporal chaotic dynamics on an invariant manifold whose coordinates are found using an autoencoder.
This approach projects the vector field of the FOM onto the invariant manifold; then, this physics-derived vector field is either corrected using dynamic data, or used as a prior that is updated with data. In both cases, the neural ordinary differential equation approach is used.
We consider simulated data from the Kuramoto--Sivashinsky and complex Ginzburg--Landau equations. Relative to the data-only approach, for scenarios of abundant data, scarce data, and even an approximate FOM (i.e.~erroneous parameter values), the hybrid approach yields substantially improved time-series predictions.
\end{abstract}


\maketitle


\begin{quotation}
Modeling complex (chaotic) systems such as weather, chemical reaction networks, and turbulent flows is challenging because these systems involve many interacting variables and the full physics are often only partially understood. Increasingly available simulation and experimental data have driven the development of data-driven reduced-order models (ROMs), which use simplified representations to efficiently forecast system behavior. To improve predictive accuracy, we develop a hybrid reduced-order modeling framework that combines data-driven learning with known physical laws. The method uses an autoencoder to identify a low-dimensional invariant manifold and employs neural ordinary differential equations to evolve the dynamics using both physics-based and data-informed corrections.
\end{quotation}

\section{Introduction} 
\label{sec:intro}

The ability to efficiently and accurately forecast complex systems is highly sought after in many applications.
\AG{Sometimes we know the full physics of these systems and can perform simulations by discretizing and solving the governing equations --- this is the case where we have an \textit{exact} ``full-order model" (FOM). Since multi-query tasks such as control or design demand the FOM to be evaluated many times, directly solving the FOM is computationally prohibitive for systems exhibiting multiscale, spatiotemporal chaos. Moreover, using the FOM to calculate adjoint models and/or Jacobians (as for \MDGrevise{multi-query} downstream tasks such as optimization or finding invariant solutions) is even more computationally intractable than simply time-integrating the FOM. Often times, however, we only have partial knowledge of the physics (e.g.~flows of complex fluids) --- this is the case where we only have an \textit{approximate} FOM, and the discussion on how fast it is to evaluate/use for high-fidelity modeling becomes moot.}
For these reasons, here we are interested in developing high-fidelity, computationally efficient reduced-order models (ROMs) of the long-time dynamics of systems with complex spatiotemporal chaotic behavior.
In this work, we combine data-driven methods with a physics-based FOM (which can be exact or approximate) to develop a hybrid ROM for forecasting spatiotemporally chaotic systems, demonstrating that using even an approximate FOM can substantially enhance predictive performance relative to an approach using only data.


Perhaps the simplest approach for hybrid data-physics ROMs for systems governed by partial differential equations, is the ``proper orthogonal decomposition (POD)-Galerkin" method \cite{Holmes_2012}. POD, also known as principal component analysis (PCA), uses data to discover an orthogonal basis ordered by variance (energy in the case of fluid flow): I.e.~the first basis vector points in the direction of maximal variance in the data set, and so on. In POD-Galerkin, the governing equations are projected onto a subset of the POD basis, yielding evolution equations for the POD amplitudes. This approach has been applied to the Navier-Stokes equations in an effort to develop low-dimensional models of turbulent flows \cite{Holmes_2012,Troy2005,Gibson}. A significant limitation of POD-Galerkin is that it implicitly takes the data to lie on a linear subspace of the full state space (i.e.~a flat manifold), which is not generally a good assumption, and may lead to a very large number of POD modes being required to yield good performance.


Nevertheless, while data from a PDE such as the Navier-Stokes Equations (NSE) may not lie in a linear subspace, it is expected at long times to lie on a finite-dimensional invariant manifold because of energy dissipation \cite{Temam1997,Zelik2022}. \MDGrevise{The Kuramoto-Sivashinsky equation\cite{Nepomnyashchii1975,Kuramoto1978, Yang2009,Ding2016,Pathak2018,Edson2019, Linot2020} is a model problem PDE in one space dimension that has been particularly well-studied in this context and is the main test-bed for the present work.}
The foundation of this work is in using simulation data to find a coordinate representation of this invariant manifold, and then finding the vector field that determines the dynamics on this manifold.
For the first step of finding a nonlinear coordinate transformation to an invariant manifold, one can use an undercomplete autoencoder \cite{Hinton2006,Goodfellow2016}. An autoencoder uses a feedforward neural network to map data from the state space to a low-dimensional latent space, and another neural network for the reverse mapping. AEs have been used to reduce dimension in turbulent flows \cite{Milano2002,Solera-Rico.2024.10.1038/s41467-024-45578-4,Omata2019} and other chaotic partial differential equation (PDE) systems \cite{Koronaki.2024.10.1016/j.jcp.2024.112910,Lee2020}. For the second step of learning the dynamics on the manifold coordinates, one can use the neural ordinary differential equations (NODEs) \cite{Chen2019} method. Using the time-dependent data, a NODE learns the vector field (\AG{right hand side, or} RHS) of the dynamical system governing the data as a neural network. The latent space coordinates found using an autoencoder can also be forecasted with recurrent neural networks (RNNs), as demonstrated in Ref.~\onlinecite{Vlachas.2022.10.1038/s42256-022-00464-w,Maulik2021}); more general approaches for reduced-order, dynamical modeling can be found in Refs.~\onlinecite{Ferguson2011,Cenedese2022,ssm_chaos}. We denote the approach of using a NODE to time-evolve the low-dimensional latent space given by an autoencoder as ``data-driven manifold dynamics" (DManD) \cite{Linot2020}. Altogether, DManD is a purely data-driven ROM. DManD has been previously used for reduced-order modeling of several chaotic dynamical systems \cite{Linot.2023.10.1016/j.jcp.2022.111838,Linot.2023.10.1016/j.ijheatfluidflow.2023.109139,Jesus.2023.10.1103/physrevfluids.8.044402,Linot.2022.10.1063/5.0069536,Young.2023.10.1103/physreve.107.034215,Floryan.2022.10.1038/s42256-022-00575-4,Zeng.2021.10.1103/physreve.104.014210,Linot:2020hu,Fox.2023.10.1103/physrevfluids.8.094401,Zeng.2022.10.1098/rspa.2022.0297}. Incorporating symmetries in DManD substantially aids in training efficiency and predictive performance \cite{zdh,DeJesus.2023.10.1103/physrevfluids.8.044402,Perez.2025.10.1103/ts3k-flx6}. Linot and Graham \cite{Linot.2023} found that with fewer than 20 degrees of freedom, a DManD model of weakly turbulent plane Couette flow quantitatively captured key features of the flow that required 1000 degrees of freedom in a POD-Galerkin approach, despite the DManD approach having no information about the underlying governing equations. The present work illustrates how to augment DManD with a FOM. 

Dimension reduction aside, a widely used method for combining physics and machine learning for learning solutions to PDEs is physics-informed neural networks (PINNs) \cite{Raissi:2019hv,Karniadakis.2021.10.1038/s42254-021-00314-5}. PINNs use a physics-based loss term that penalizes the residual of the PDE at collocation points to learn PDE solutions that are physically consistent.
While PINNs may also include a supervised loss on the initial and boundary conditions from data measurements, the solution is mostly influenced by the purely physics-based loss --- therefore, PINNs are often regarded as ODE or PDE solvers that are not data-driven.
Nonetheless, the use of a PINN-like loss term in data-driven frameworks is quite common. One example is in physics-informed deep learning \cite{Raissi2017}, where the data-driven loss is placed on points outside of the initial and boundary conditions. in Refs.~\onlinecite{Wang2021,Li2024}, a PINN-like loss term is used with neural operators \cite{Lu2021,Li2020}, which are data-driven techniques for learning PDE solution operators.
For the most part, PINNs are not designed to produce accurate solutions outside the time horizon of the collocation points used in training, though some work \cite{Kim2020} has been done to address this issue.
By contrast, predicting forward in time over an arbitrarily long horizon is the central topic of the present work.
Because PINNs are not necessarily data-driven, not reduced-order, and not designed for time-series prediction, we do not consider this approach further.

We now arrive at the physics-based method on which our hybrid ROM described later is based. This method is called Manifold Galerkin (MG) \cite{Lee2020}. MG is an accurate FOM-based method that nonlinearly projects the governing equations onto a manifold learned from data in the form of an autoencoder. That is, Manifold Galerkin is similar to POD-Galerkin, just with a nonlinear rather than linear coordinate transformation to a lower-dimensional manifold. The approach of using MG assumes an exact FOM is known.
Although the MG model provides the vector field in a reduced-dimensional space (which can then be time-integrated for trajectory predictions), it is often more expensive to integrate than the FOM because the vector field must still be evaluated in the original high-dimensional state space at each time step.
The authors of Ref.~\onlinecite{Lee2020} suggest hyper-reduction approaches (i.e.~reducing the mesh size on which the FOM is evaluated) to address this problem of MG, which has been demonstrated in Ref.~\onlinecite{Romor2023}. MG will be discussed in more detail in Section \ref{sec:mg}.

To the best of our knowledge, Ref.~\onlinecite{Sholokhov2023} is the only work besides ours that uses MG as a physics-based method to inform a data-driven forecasting model. In that work, the authors developed a hybrid physics-informed NODE (PINODE) method, which applies a soft constraint to the vector field on an autoencoder latent space learned by a NODE; this soft constraint drives the RHS learned by the NODE to the RHS of the MG model, where the former is more efficient to integrate than the latter. Our hybrid ROM (described next) differs from the PINODE method \cite{Sholokhov2023} in a few key ways, which are described in Section \ref{sec:pi_dmand}. Another hybrid method is presented in Ref.~\onlinecite{Pathak2018}, where the FOM (subject to some approximation error) is directly combined with a reservoir computing network to forecast the state space of a chaotic system; however, since reservoir networks have a hidden state with dimensionality higher than the original state space, this hybrid method is not a ROM.
Additionally,
reservoir networks are inherently non-Markovian techniques (which is also true for RNNs \cite{Vlachas2018a,Vlachas2019} and transformers \cite{Wen2023,Gilpin2023,Solera-Rico.2024.10.1038/s41467-024-45578-4}) 
that use past history information to make time series predictions. For data that comes from a memoryless dynamical system such as differential equation, these approaches do not maintain this dynamical systems structure. Our aim is to develop reduced-order models that maintain a differential equation structure.

In this paper, we combine the purely data-driven approach DManD with the physics-based MG method to develop a hybrid ROM framework that we call physics-informed DManD (PI-DManD)\MDGrevise{, and demonstrate proof-of-concept for the method with two widely-studied model problems}. We describe our methods for constructing the hybrid ROM in Section \ref{sec:methods}. In Section \ref{sec:res}, we apply the approach to the prediction of spatiotemporal chaos in the Kuramoto--Sivashinsky equation when varying the amount of training data and the accuracy of the physics-based prior information provided by MG. Appendix \ref{sec:cgle} includes results for the complex Ginzburg--Landau equation (CGLE) \cite{Aranson2002,Shraiman1992}, another system exhibiting spatiotemporal chaos. In all cases, PI-DManD leads to substantial improvements in predictive capability relative to the data-only DManD approach.

\section{Methods}
\label{sec:methods}

We work with state space data $u \in \mathbb R^{d_u}$, where $d_u$ is the state space dimension, sampled with uniform data spacing $\tau$. Given time series data of $u$ generated from a PDE or experiments, we aim to develop a ROM that can be used for time evolution predictions. Besides reducing the number of degrees of freedom in the system, we desire the ROM to give accurate short-time predictions and long-time predictions that capture key statistical features of the true attractor.

Here, we use a single long time series of $u$ as the dataset. Initial transients are removed, so the data lies on an attractor that is embedded in an invariant manifold of the underlying dynamical system. We denote this dataset as $\{u_i\}_{i=1}^M$, where $u_i=u(t_i)$ is the snapshot of the state at time instant $t_i$ and $M$ is the number of snapshots. The time series data come from the underlying (ground truth) autonomous dynamical system
\begin{equation}
    \frac{du}{dt}=f(u;\mu) ,
    \label{eq:FOM}
\end{equation}
where $f$ is the true vector field that depends on physical parameters $\mu$ (e.g.~domain length). Unless the data are from simulations, $f$ and $\mu$ are generally not known. Still, one often has some knowledge or model of the underlying physics. Assume that we know a full-order model (FOM) of the true dynamical system in Equation \ref{eq:FOM},  written as
\begin{equation}
    \frac{du}{dt}=f_\FOM(u;\mu_\FOM) ,
    \label{eq:FOM_approx}
\end{equation}
where $f_\FOM$ is the vector field for the model and $\mu_\FOM$ are the physical parameters of the FOM. The field $f_\FOM$ may only be approximate due to the lack of full first-principles knowledge of the system, while $\mu_\FOM$ may only be approximate if the data are from experiments and there is uncertainty in the parameters. 

In Section \ref{sec:dmand}, we start by describing a solely data-driven ROM. Then in Section \ref{sec:mg}, we discuss how the MG method \cite{Lee2020} uses the FOM in Equation \ref{eq:FOM_approx} to create a physics-based time evolution method. Lastly, we present our hybrid ROM in Section \ref{sec:pi_dmand}, which uses the MG method to incorporate physics knowledge, while also using data to correct for the FOM approximation errors that are not addressed by MG.

\subsection{Data-driven manifold dynamics (DManD)}
\label{sec:dmand}

We first find a low-dimensional, high-fidelity latent representation of $u$ for which to build the ROM. To do so, we will use an undercomplete autoencoder \cite{Hinton2003} to discover a coordinate transformation to (and from) a low-dimensional latent space representation $h \in \mathbb R^{d_h}$, where $d_h < d_u$. As mentioned in the Introduction, dissipative systems often have long-time dynamics that lie on an invariant manifold with dimension $d_\mathcal M \ll d_u$. In the present work, we assume that this is the case for the systems under consideration, and further hypothesize that points on the attractor suffice to train an autoencoder that yields a good approximate coordinate representation of the invariant manifold in the neighborhood of the attractor. Indeed, the quality of the autoencoder is a central issue in the performance of the methods we describe. 

While it is generally difficult to determine $d_\mathcal M$, there are a variety of methods for estimating it \cite{Ding2016,irmae-c} (see Appendix \ref{sec:cgle_dm} for a discussion of one such method and how we applied it to the CGLE).
\AG{Since we consider data $u_i$ living on an attractor that is embedded in an invariant manifold, an autoencoder can in principle be trained to find a latent space whose components represent the coordinates on this manifold. We note that finding the coordinates on the invariant manifold and learning the dynamics on it can become complicated with the presence of noise, though some work \cite{Tran2024} has been done to address this challenge.}

An autoencoder consists of two back-to-back neural networks. The encoder $\Enc$ nonlinearly maps $u_i$ to $h_i = \Enc(u_i)$. The decoder $\Dec$ performs the reverse mapping $\tilde u_i = \Dec(h_i)$, where $\tilde u_i$ is the reconstructed snapshot. We minimize the reconstruction error of the autoencoder (AE) with the loss function
\begin{equation}
    \mathcal L_\mathrm{AE} = \langle \|u_i - \Dec(\Enc(u_i; \theta_\Enc); \theta_\Dec) \|_2^2 \rangle_i ,
    \label{eq:AE}
\end{equation}
where $\theta_\Enc$ and $\theta_\Dec$ are, respectively, the neural network parameters of the encoder and decoder networks and $\langle \cdot \rangle_i$ denotes an ensemble average over a batch of training snapshots. Here, and elsewhere, we use stochastic gradient descent to minimize the loss function.

With the autoencoder in hand, the DManD approach to reduced-order modeling learns a dynamical system in the latent space:
\begin{equation}
    \frac{dh}{dt}=g(h) ,
    \label{eq:ROM}
\end{equation}
where $g$ is the true vector field in the manifold coordinates. We train a neural ODE (NODE) \cite{chen2018neural} to approximate this function. Given a latent state $h(t_i)$, one can calculate
\begin{equation}
    \hat h_\mathrm{DManD}(t_i+\tau) = h(t_i) + \int_{t_i}^{t_i+\tau} g_d(h(t'); \theta_{g_d}) \, \mathrm d t' ,
    \label{eq:ROM_int}
\end{equation}
where $\hat h_\mathrm{DManD}(t_i+\tau)$ is the NODE prediction of the latent state at the next sampled time instant, and $g_d$ is the data-driven approximation of $g$. We model $g_d$ with a deep feedforward neural network and optimize its parameters $\theta_{g_d}$ with the loss function 
\begin{equation}
    \mathcal L_\mathrm{DManD} = \langle \| \hat h_\mathrm{DManD}(t_i+\tau) - h(t_i+\tau) \|_2^2 \rangle_i .
\end{equation}
Here, we only consider the reconstruction loss for an integration time horizon spanning a single data spacing interval $\tau$. While one could consider a multi-step loss with a longer time horizon, we find that, for the systems studied here, doing so does not improve the results and slows down training. Nevertheless, the use of a multi-step NODE loss may be helpful in some situations, such as for time-delay embeddings of partially observed data \cite{Young.2023.10.1103/physreve.107.034215}. In the results, we will increase the data spacing $\tau$ used in the NODE training. 
Lastly, we note that it has been recently shown \cite{Buzhardt2025} that the NODE approach is equivalent to a Koopman-operator-based method called extended dynamic mode decomposition with dictionary learning (EDMD-DL) \cite{Kevrekidis.2017} with a projection step onto the state.

\begin{figure*}
	\centering
	\includegraphics[width=1\linewidth]{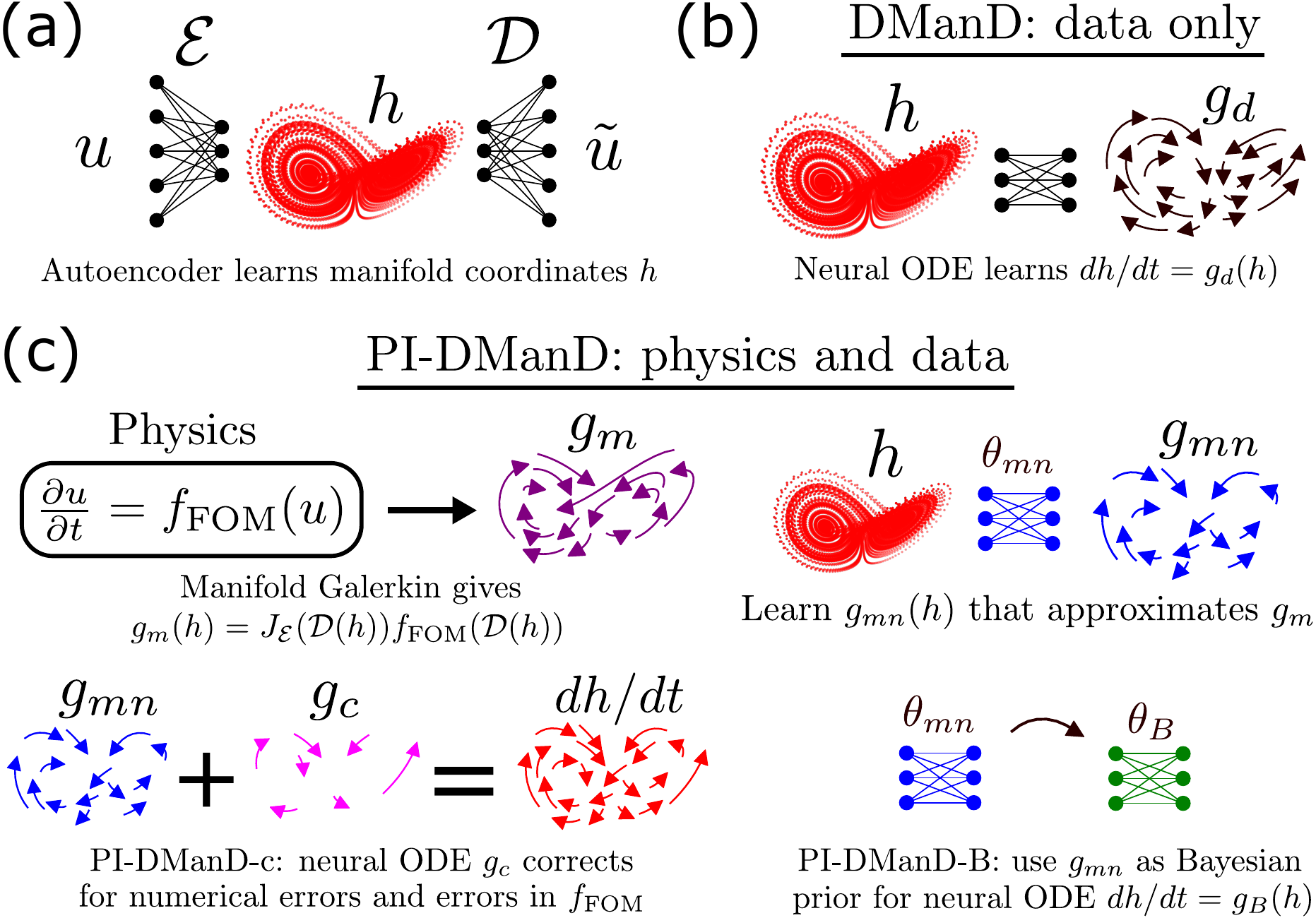}
    \caption{Data-driven manifold dynamics (DManD) is a data-driven ROM that reduces dimension of the state $u$ with an autoencoder (a) and evolves trajectories with a vector field on the latent space $h$ learned by a neural ODE (b). (c) Physics-informed DManD (PI-DManD) is a hybrid ROM that uses Manifold Galerkin to project the vector field on the full order model (FOM) $f_\FOM$ onto that of the latent space, learns a neural network approximation of the latent space vector field as $g_{mn}$, and either uses a corrective neural ODE $g_c$ to fix numerical and FOM approximation errors (PI-DManD-c) or uses $g_{mn}$ as a Bayesian-motivated prior for training a DManD model (PI-DManD-B).}
    \label{fig:sketch}
\end{figure*}

Overall, this combination of an autoencoder and a NODE to build a ROM is coined data-driven manifold dynamics (DManD) \cite{Linot2020}. A schematic of DManD is depicted in Figure \ref{fig:sketch}(a) and \ref{fig:sketch}(b).
As a side note, NODE models can be stabilized for long-time predictions by following \cite{Linot2022}, in which the integrand $g_d(h;\theta_{g_d})$ in Equation \ref{eq:ROM_int} is modified to $g_d(h;\theta_{g_d}) -ah$, where $a>0$. Although \cite{Linot2022} found that adding the linear damping term $ah$ primarily benefits the case where there is no dimension reduction (i.e.~a NODE is used to approximate $f_\FOM$), we include the damping term nonetheless when learning $g_d$ since doing so also has no negative effects (we use $a=10^{-3}$ for this paper).
\AG{The size of the damping term does not affect the performance of the ROM as long as it is a few orders of magnitude smaller than the magnitude of the data (order 1 here).}
Lastly, we note that it is possible to simultaneously learn a latent space and a NODE model; however, we find that jointly training an autoencoder and a NODE model often requires more expressive networks, leads to slower training convergence, and does not yield improved performance.

\subsection{Manifold Galerkin}
\label{sec:mg}

We now present the formalism for the Manifold Galerkin (MG) method proposed in Ref.~\onlinecite{Lee2020}, which directly expresses the vector field $f_\FOM$ in the manifold coordinates $h$.
\MDGrevise{We assume for the moment that the autoencoder determines an exact coordinate transformation between the latent and ambient representations of the data.}
Using the chain rule and substituting $u = \Dec(h; \theta_\Dec)$ into the FOM in Equation \ref{eq:FOM_approx} yields
\begin{equation}
    \frac{d h}{dt} = \frac{d \Enc(u)}{du} \frac{d u}{dt} = \frac{d \Enc(u)}{du} f_\FOM(\Dec(h; \theta_\Dec); \mu_\FOM) ,
    \label{eq:chain_rule}
\end{equation}
Let the Jacobian of the coordinate transformation between $u$ and $h$ be expressed as
\begin{equation}
	\frac{\partial \Enc(u)}{\partial u} = J_\Enc(\Dec(h; \theta_\Dec);\theta_\Enc) .
\end{equation}
Since $\Enc(u)$ is a neural network, automatic differentiation can be used to easily compute $J_\Enc$. Substituting this expression for $J_\Enc$ into Equation \ref{eq:chain_rule} yields the final equation for MG:
\begin{equation}
\begin{aligned}
       \frac{d h}{dt} &= g_m(h;\theta_\Enc, \theta_\Dec, \mu_\FOM), \\
    g_m &=J_\Enc(\Dec(h; \theta_\Dec);\theta_\Enc) \, f_\FOM(\Dec(h; \theta_\Dec); \mu_\FOM).
    \label{eq:MG}
\end{aligned}
\end{equation}
\AG{As discussed in Section \ref{sec:dmand}, the autoencoder is trained on data living on the invariant manifold. As a result, assuming the FOM and its parameters, \MDGrevise{as well as the autoencoder,} are exact (i.e.~$f_\FOM=f$ and $\mu_\FOM=\mu$), the full vector field is tangent to the invariant manifold coordinates found by the autoencoder. Then, we also have that $g_m(h)$ given by MG is the true vector field on the invariant manifold, expressed in the manifold coordinates $h$.}  \MDGrevise{Beck and Schneider\cite{Beck20261048550arxiv260101590} used MG, with an additional autoencoder loss term to weakly enforce tangency between the autoencoder-approximated invariant manifold and the vector field, to compute recurrent solutions such as periodic orbits for the KSE and a model Navier-Stokes system, the Kolmogorov flow.}



Although MG in Equation \ref{eq:MG} has the appearance of a ROM since it provides the vector field in the latent space $h$, it is in fact generally slower to solve than the FOM. At each time step, MG not only retains the expense of computing $f_\FOM$, but must also first decode $h$ to $u$ and evaluate a Jacobian (which, for $J_\Enc$, involves another encoding step).
We quantify the speed of the MG method in Appendix \ref{sec:kse_train_speed}.
Work has been done to accelerate the MG method using hyper-reduction techniques \cite{Romor2023}, which aim to perform approximate FOM computations that are independent of the ambient dimension $d_u$.
\AGrevise{Overall, MG is a physics-based reduced-order model that projects the FOM vector field onto the manifold coordinates found with a data-driven autoencoder. However, because evaluating MG requires the FOM and encoder Jacobian at every time step, we do not consider it a ROM in the computational sense used here, where a ROM should reduce the cost of repeated model evaluations.}

The run-time expense of MG can be overcome by using 
a neural network to learn an approximation of the function $g_m(h)$. Since we know the functional form of $f_\FOM(u)$, we can use the data points $u_i$ to generate $f_{\FOM,i}=f_\FOM(u_i)$. If the data come from simulations of the FOM \AGrevise{that directly calculate the RHS function values at the states corresponding to saved snapshots (for example, an explicit time stepping method)}, then $f_{\FOM,i}$ is already computed at each time step of the simulation, so we can store it instead of recomputing it. Otherwise, the step to generate $f_{\FOM,i}$ from $u_i$ for $i=1,\dots, N$ only needs to be performed once as a preprocessing step. We then calculate the MG output data $g_{m,i} = J_\Enc(u_i) f_{\FOM,i}$ for all $i$. We denote the neural network for mapping the inputs $h_i=\Enc(u_i)$ to the MG outputs $g_{m,i}$ as $g_{mn}$, and we train it with the loss
\begin{equation}
    \mathcal L_\MGN = \langle \| g_{mn}(h_i; \theta_{g_{mn}}) - g_{m,i}) \|_2^2 \rangle_i .
\end{equation}
Here, only the parameters $\theta_{g_{mn}}$ are learnable (i.e, $g_{m,i}$ are constants and do not depend on the network parameters of the autoencoder). One can substitute $g_m(h)$ in Equation \ref{eq:MG} with $g_{mn}(h)$ to get
\begin{equation}
    \frac{d h}{dt} = g_{mn}(h; \theta_{g_{mn}}) .
    \label{eq:MGN}
\end{equation}
Since $g_{mn}(h; \theta_{g_{mn}})$ is a neural network approximation of $g_m(h)$, we call this method MG - neural network (MG-N). As will be shown in Section \ref{sec:res}, the MG-N model in Equation \ref{eq:MGN} is much faster to integrate than the MG model in Equation \ref{eq:MG}. 
Note that the dynamic nature of the data set is not used in development of MG/MG-N, so if the autoencoder is accurate, but the FOM is not, these approaches will model an incorrect vector field, albeit on a correct set of manifold coordinates.

\subsection{Physics-informed data-driven manifold dynamics (PI-DManD)}
\label{sec:pi_dmand}

Now we describe how we integrate MG-N with DManD to create two physics-informed DManD (PI-DManD) methods. Figure \ref{fig:sketch}(c) shows a schematic of these approaches.
When using MG-N alone as a ROM, there are three sources of errors: numerical and test error in the autoencoder that influence the training of $g_{mn}$, test error in $g_{mn}(h)$ itself, and lastly, errors in the FOM. The last type of error grows as the physics-based approximation of the physics $f_\FOM(u,\mu_\FOM)$ deviates further from the true underlying physics $f(u,\mu)$ governing the data. FOM error can occur if $f_\FOM \neq f$ and/or $\mu_\FOM \neq \mu$; the latter case may occur if there is uncertainty in the physical or PDE parameters (e.g.~domain length), which will often arise if the dataset comes from experiments.

In our first PI-DManD approach, denoted PI-DManD-c, we account for all three types of errors by adding a correction term, $g_c(h)$, to the vector field from MG-N to get our final, reduced-order dynamical model in the latent coordinates:
\begin{equation}
    \frac{dh}{dt}=g_{mn}(h; \theta_{g_{mn}}) + g_c(h; \theta_{g_c}) .
    \label{eq:pidmand}
\end{equation}
We learn $g_c(h; \theta_{g_c})$ using a NODE approach with loss
\begin{equation}
    \mathcal L_\text{PI-DManD-c} = \langle \| \hat h_\text{PI-DManD-c}(t_i+\tau) - h(t_i+\tau) \|_2^2 \rangle_i ,
\end{equation}
where
\begin{equation}
    \begin{gathered}
    \hat h_\text{PI-DManD-c}(t_i+\tau) = h(t_i) + \\
    \int_{t_i}^{t_i+\tau} \left[ g_{mn}(h(t'); \theta_{g_{mn}}) + g_c(h(t'); \theta_{g_c}) \right] \, \mathrm d t' .
    \end{gathered}
\end{equation}
Here, only the weights $\theta_{g_c}$ are trainable (the MG-N weights $\theta_{g_{mn}}$ are kept frozen). Altogether, PI-DManD-c learns a function $g_{mn}+g_c$ for the latent space dynamics. While MG-N is a physics-informed ROM that can be heavily biased by the FOM approximation, PI-DManD-c is a hybrid ROM that is less susceptible to this bias, as will be illustrated in the results later.
\AG{Note, if the physics-based information is accurate, the MG-N vector field $g_{mn}$ will be close to $g$ (the exact vector field in the latent space determined from data) and the difference $g - g_{mn} = g_c$ will be small; also, in the limit of an exact $g_{mn}$ (and no numerical errors with the autoencoder or MG-N), NODE training with weight decay will lead $g_c$ to vanish. If the physics-based information is not accurate, PI-DManD-c will just learn a larger correction $g_c$ --- given that the information encoded in $g_{mn}$ would still be relevant as the NODE for $g_c$ is being trained in this case, PI-DManD-c will not simply reduce to DManD.}

The second PI-DManD approach, denoted PI-DManD-B, uses the MG-N model as a prior for training the NODE model.
Specifically, we aim to learn a NODE model for the RHS of
\begin{equation}
    \frac{dh}{dt}=g_B(h; \theta_{g_B}) ,
    \label{eq:pi_dmand_b}
\end{equation}
with the loss
\begin{equation}
    \begin{gathered}
    \mathcal L_\text{PI-DManD-B} = \langle \| \hat h_\text{PI-DManD-B}(t_i+\tau) - h(t_i+\tau) \|_2^2 \rangle_i \\
    + \lambda_B \| \theta_{g_B} -\theta_{g_{mn}} \|_2^2 ,
    \label{eq:pi_dmand_b_loss}
    \end{gathered}
\end{equation}
where
\begin{equation}
    \hat h_\text{PI-DManD-B}(t_i+\tau) = h(t_i) + \int_{t_i}^{t_i+\tau} g_B(h(t'); \theta_{g_B}) \, \mathrm d t' .
\end{equation}
The difference here as compared to a DManD model lies in the additional regularization term $\| \theta_{g_B} -\theta_{g_{mn}} \|_2$, which acts to penalize deviations of the NODE weights from the MG-N weights. The prefactor $\lambda_B$ controls the strength of the regularization. By using this regularization term, the NODE model learns the vector field from data while also being softly constrained to the physics-derived vector field from MG-N --- as compared to the correction NODE used in PI-DManD-c, this NODE for $g_B$ is more natural in learning from the data and physics simultaneously, which yields benefits as will be shown later. Another difference is that here the NODE $g_B$ must have the same architecture as the MG-N network, which is mostly a minor restriction.
In addition to the regularization term, we perform a
\AGrevise{warm start by} initializing the weights of $g_B$ to those of MG-N. So, the PI-DManD-B model uses the MG-N model $g_{mn}$ as a physics-based prior.

The loss used by PI-DManD-B can be viewed as being motivated from a Bayesian framework. To see this, consider the regression problem $y = m(x;\theta_m)$, where $\theta_m$ are the model parameters. According to Bayes' theorem, we can write the posterior distribution for the parameters $\theta_M$ given the data $y$ as
\begin{equation}
    p(\theta_m | y) = \frac{p(y | \theta_m) p(\theta_m)}{p(y)} .
    \label{eq:bayes}
\end{equation}
Assuming a normal distribution for the likelihood $p(y | \theta_m) \propto \exp(-(y-m)^2/2\sigma_{m}^2)$ as well as for the prior $p(\theta_m) \propto \exp(-(\theta_m-\theta_p)^2/2\sigma_p^2)$, maximizing the log likelihood corresponds to minimizing the loss function
\begin{equation}
    \mathcal L = \frac{1}{2 \sigma_{m}^2} \langle (y - m)^2 \rangle + \frac{1}{2 \sigma_p^2} \| \theta_m - \theta_p \|_2^2.
    \label{eq:bayes_loss}
\end{equation}
This expression is very similar to Eq.~\ref{eq:pi_dmand_b_loss}, motivating the description of PI-DManD-B as a Bayesian approach.
\AG{Nevertheless, we emphasize that while its loss function can be motivated from Bayes' theorem, the PI-DManD-B model is deterministic and we are not employing a Bayesian inference framework.}

It must be noted that PI-DManD bears similarities to the PINODE method of Ref~\onlinecite{Sholokhov2023}. PI-DManD and PINODE use knowledge of the governing equations to get the physics-derived vector field on the latent space and combines that with a NODE; however, our method differs in several key ways. First, PI-DManD only requires a dataset composed of a time series of data points $u_i$, while PINODE requires additional data points called ``collocation points'' that are user-specified.
\AG{Second, PI-DManD calculates vector field information $f_{\FOM,i}$ only from the state $u_i$ (which is the only data we work with), while PINODE calculates $f_{\FOM,i}$ only from the user-specified collocation points and not the state data $u_i$.}
Third, PI-DManD front-loads the FOM evaluations required by MG into a preprocessing step before training the intermediate MG-N neural network, whereas PINODE performs these generally expensive FOM evaluations during their NODE training --- as a result, PI-DManD would generally train faster than PINODE.
\AG{Fourth, unlike PINODE, we will later consider a case where only an approximate FOM is available for supplying the physics-based information.}
Lastly, PINODE was only applied to systems with relatively simple long-time dynamics; systems with substantially higher complexity are considered here.


\AG{PI-DManD builds a ROM solely from timeseries data of the state $u_i$ (from experiments or simulations) supplemented with information from a FOM (which may or may not be exact). In general, PI-DManD is intrusive in the sense that code modification may be required to calculate the vector field $f_{\FOM,i}$ from $u_i$ before proceeding with MG/MG-N. One could consider the case where no FOM is known; then, if the state data is closely spaced and has low noise, one could take numerical derivatives of the state space data to estimate the vector field $f_{\FOM,i}$ and proceed with training a PI-DManD model. However, while this approach is non-intrusive, it is not generally applicable because of the limitations just mentioned. In our previous work \cite{alec_chaos} and this work (see Section \ref{sec:scarce_data_kse}), we show that DManD/PI-DManD works well even when the data spacing is half of a Lyapunov time.
}

\subsection{Architecture and training details}
\label{sec:train_details}

A summary of the architectures used for all neural networks introduced in Section \ref{sec:methods} is provided in Table \ref{table:arch}. Training is performed in PyTorch \cite{Paszke2019}. The first 80\% of the times series is used for training, and the remaining 20\% for testing. For all neural networks, we use an AdamW optimizer with a multistep learning rate scheduler that starts the learning rate at $10^{-3}$ and halves it 7 times until $\approx 10^{-5}$. By default, we use a weight decay of $10^{-4}$; however, for the networks trained in the scarce data case in Section \ref{sec:scarce_data_kse} we increase the weight decay strength to a value within the range of $10^{-3}$ to $10^0$ (a sweep is done to decide on the value).

\begin{table*}
\renewcommand{\arraystretch}{1.5}
\centering
\begin{tabular}{ c|c|c|c } 
     Network & Function & Layers & Activations \\
     \hline
     Encoder & $\Enc$ & $d_u$:4$d_u$:2$d_u$:$d_h$ & ReLU:ReLU:Linear \\
     Decoder & $\Dec$ & $d_h$:2$d_u$:4$d_u$:$d_u$ & ReLU:ReLU:Linear \\
     NODE (DManD) & $g_d$ & $d_h$:256:256:$d_h$ & GeLU:GeLU:Linear \\
     Neural network (MG-N) & $g_{mn}$ & $d_h$:256:256:$d_h$ & GeLU:GeLU:Linear \\
     NODE (PI-DManD-c) & $g_c$ & $d_h$:256:256:$d_h$ & GeLU:GeLU:Linear \\
     NODE (PI-DManD-B) & $g_B$& $d_h$:256:256:$d_h$ & GeLU:GeLU:Linear
\end{tabular}
\caption{Neural network architectures for the autoencoder, DManD, MG-N, and the two PI-DManD methods.}
\label{table:arch}
\end{table*}

For the autoencoder used to learn $\Enc$ and decoder $\Dec$ (see architectures in Table \ref{table:arch}), the dimension of the latent space $h$ needs to be determined. For the KSE system considered in this paper, the dimension of the invariant manifold at $L=22$ is known to be $d_\mathcal M=8$ \cite{Linot2020,Linot2021,irmae-c,Ding2016}. 
For the CGLE system considered in Appendix \ref{sec:cgle}, we estimate the dimension of the invariant manifold using an implicit rank-minimizing autoencoder \cite{jing2020implicit,irmae-c} (see Appendix \ref{sec:cgle_dm} for details). 
For the autoencoder, we use a batch size of 64 and train for 2000 epochs. For the neural network used to learn $g_{mn}(h)$ for MG-N (see architecture in Table \ref{table:arch}), we use a batch size of 256 and train for 2000 epochs.

The training details for the NODEs $g_d$, $g_c$, and $g_B$ (see architectures in Table \ref{table:arch}) are as follows. We use the 5th order Dormand-Prince (dopri5) adaptive time-step integrator in the \texttt{torchdiffeq} package \cite{chen2019neural} with default tolerances. For all NODE training, we use a batch size of 64 and train for 100,000 \textit{iterations}. While the terms ``iterations'' and ``epochs'' are often used interchangeably in the research community with regards to NODE training, the former is the nomenclature used in the original NODE paper \cite{chen2018neural}. In an iteration, weights are updated by randomly sampling a single batch (size 64 in our case) from the entire dataset and performing a forward and backward pass, while a single epoch performs many iterations until the entire dataset is sampled. The published NODE code in Ref.~\onlinecite{chen2018neural} uses iterations, so we do too in our adapted code. For training the NODE model correction $g_c$ for PI-DManD-c in particular, we use near-zero weight initialization (all other neural network training is with Pytorch's default Xavier weight initialization); this choice is motivated by the fact that $g_c$ would be zero if $g_{mn}$ in Equation \ref{eq:pidmand} were exact (i.e.~if there were no FOM approximation error). Lastly, for PI-DManD-B, we use $\lambda_B=10^{-5}$ (see Equation \ref{eq:pi_dmand_b_loss}) for the strength of the regularization parameter. For the systems considered in this paper, we found that the results (even if the FOM is not exact) are robust to the value of $\lambda_B$ as long as it is less than $10^{-3}$; in the limit of $\lambda_B=0$, where PI-DManD-B is still robust, the method just involves the \AGrevise{warm start} of initializing the NODE weights to those of the physics-based prior (which suggests that the main advantage of PI-DManD-B comes from the initialization).

\section{Results} 
\label{sec:res}

We work with data generated from the Kuramoto--Sivashinsky equation (KSE), which is a 1D PDE exhibiting spatiotemporal chaos and is widely used as a test bed for data-driven methods \cite{Yang2009,Linot2020,Pathak2018}. \AG{The KSE has been used to model physical systems such as the instabilities in flame fronts and thin liquid films \cite{Kuramoto1978,Nepomnyashchii1975},} and is given by
\begin{equation}
    \frac{\partial u}{\partial t} = -u \frac{\partial u}{\partial x} - \frac{\partial^2 u}{\partial x^2} - \frac{\partial^4 u}{\partial x^4} .
    \label{eq:kse}
\end{equation}
We use a domain length of $L=22$ with periodic boundary conditions, wherein the solution exhibits chaotic dynamics and has a Lyapunov time $\tau_L$ of 22.2 time units \cite{Ding2016,Edson2019}. For this KSE system, the dimension of the invariant manifold is empirically known to be $d_\mathcal M=8$ \cite{Linot2020,Linot2021,irmae-c,Ding2016}; therefore, we will use $d_h=8$ as the dimension for the autoencoder. We generate a long trajectory using an exponential time-differencing RK4 (ETD-RK4) solver \cite{Kassam2005} in the Fourier domain. We remove the transient and use $M = 500,000$ snapshots of $u$ discretized on $d_u=64$ uniformly spaced grid points, separated by $\tau=0.25$ time units (though we will vary this spacing later for some cases). We use the first 80\% of the dataset for training, so that the number of training snapshots $N$ is 400,000. The last 100,000 snapshots are for testing (even when we do not use the whole training dataset, as in Section \ref{sec:scarce_data_kse}).
\AGrevise{For each neural network, we train and use only a single model.}
See Section \ref{sec:train_details} for other training details.

We test whether the hybrid ROMs (PI-DManD-c and PI-DManD-B) can outperform the DManD ROM, which is solely data-driven. In Section \ref{sec:kse_lots_data}, we train the methods using the whole time series. In Section \ref{sec:scarce_data_kse}, we consider three subcases of using a subset of the time series (note, for one subcase, we will generate and use a new time series dataset with $N = 400,000$ and $\tau > 0.25$). In Section \ref{sec:kse_err_FOM}, we consider the case when the parameters used in the FOM do not match the simulation parameters used to generate the data --- this models the situation when the FOM is not exact.
(In Appendix \ref{sec:cgle}, we repeat the tests reported here for the 1D CGLE.)

\subsection{Abundant data, exact FOM}
\label{sec:kse_lots_data}

\begin{figure*}
	\centering
	\includegraphics[width=1\linewidth]{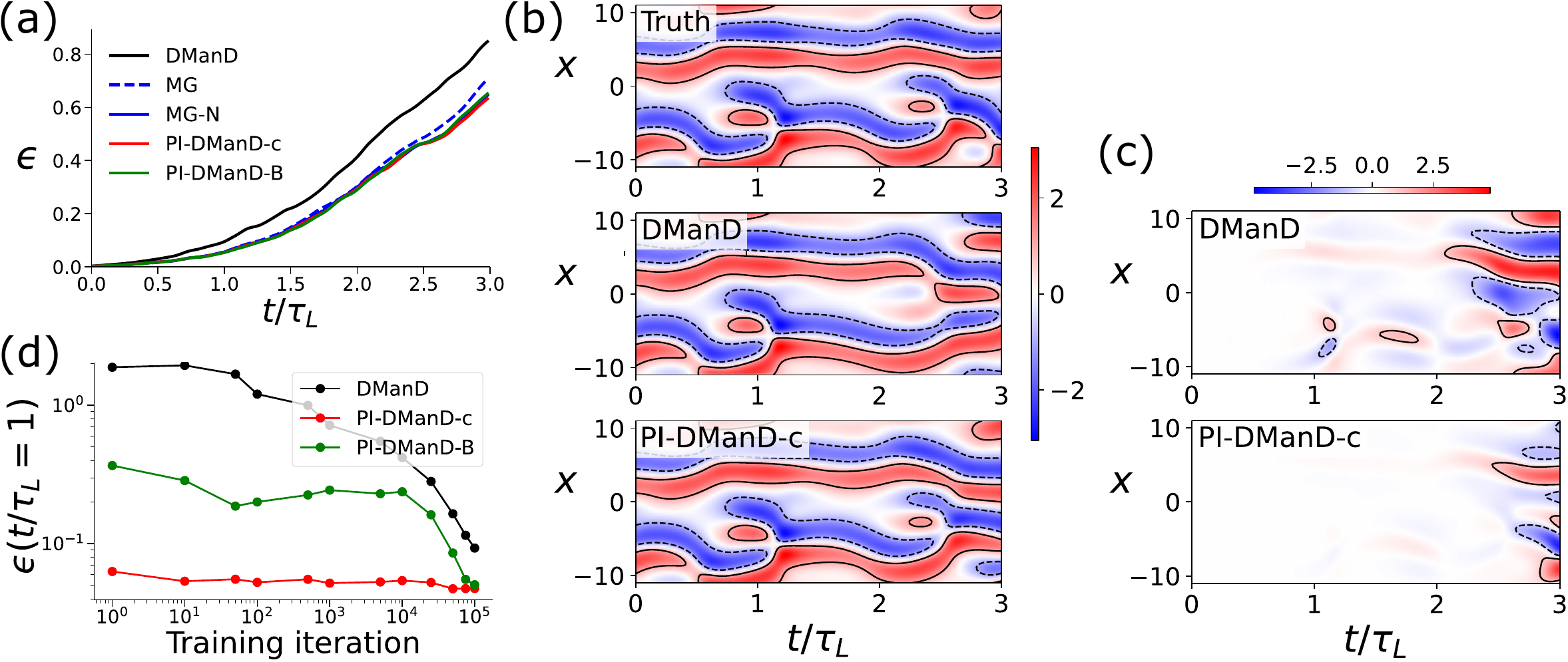}
	\caption{KSE, $L=22$ (for both data and FOM), $N=400,000$, $\tau=0.25$. (a) Ensemble-averaged short-time tracking error for all methods. (b) Representative trajectories of ground truth and model predictions (contours at $u = \pm 1$). (c) Difference plots between ground truth and model predictions (contours at difference values of $\pm 1$). Representative trajectories and difference plots for other models look similar to those of PI-DManD-c. (d) NODE training convergence for purely data-driven and hybrid methods \AGrevise{(note the log-log scale)}.}
	\label{fig:kse_res}
\end{figure*}

Here, we use the entire training dataset and the exact physics ($f_\FOM=f$) for the MG/MG-N methods. We gauge model performance on how well the model predictions (latent space trajectories that have been decoded back to the full space) track the true solution over a time horizon of $3\tau_L$ --- any good method should be able to make reasonably accurate predictions over at least $\tau_L$. We calculate the ensemble-averaged, normalized tracking error
\begin{equation}
    \epsilon(t) = \left\langle \frac{\| u(t) - \hat u(t) \|_2}{\| u(t) \|_2} \right\rangle ,
    \label{eq:eps_node}
\end{equation}
where $\hat u(t)$ is the model prediction trajectory, and the ensemble average $\langle \cdot \rangle$ is calculated using 300 non-overlapping trajectories whose initial conditions are from the testing dataset. In Figure \ref{fig:kse_res}(a), we plot the tracking error $\epsilon(t)$ for DManD, the physics-informed methods (MG and MG-N), and the two hybrid methods. Here, all non-DManD methods perform similarly and display much smaller error than DManD --- at $t=\tau_L$, their error is smaller by a factor of two; this result suggests that including physics aids in model predictions when the provided physics are exact. Figure \ref{fig:kse_res}(b) shows space-time plots of representative trajectories for ground truth and the DManD and PI-DManD-c model predictions. The differences between the ground truth and model trajectories in Figure \ref{fig:kse_res}(b) are plotted in Figure \ref{fig:kse_res}(c), indicating that noticeable differences arise sooner for DManD than for PI-DManD-c. (Representative trajectories and difference plots for other models, not shown, look similar to those of PI-DManD-c due to similar error curves $\epsilon(t)$ --- see Figure \ref{fig:kse_res}(a).) In this case where we have a large amount of closely spaced training data and the FOM is exact, all physics-informed and hybrid methods outperform DManD in short-time forecasting.

We now turn to the time used to train and evaluate the ROMs. Figure \ref{fig:kse_res}(d) shows the evolution of $\epsilon(t=\tau_L)$ during the course of NODE training, for DManD and the two PI-DManD variants. Most notably,
the correction vector field $g_c$ learned by PI-DManD-c converges several orders of magnitude faster than the vector field $g_d$ learned by DManD. \MDGrevise{With an exact model and accurate autoencoder, the MG-N model is highly accurate without any correction.} However, since PI-DManD-c (and PI-DManD-B) are derived from MG-N, the MG-N training time to learn $g_{mn}$ must also be accounted for --- in this case, the time to just train MG-N is around 4 times longer than the time to train DManD (see details in Appendix \ref{sec:kse_train_speed}).
While the combined training time for PI-DManD-c is longer than for DManD, it still has the benefit of better predictive performance. As for the inference time, all models perform comparably (a few seconds) except MG --- since MG is not a ROM, it takes 9 to 48 times longer to evolve an initial condition than the other methods (see Tables \ref{table:speed} and \ref{table:speed_cgle} in the Appendix).
\AG{Although neural network-based ROMs require a relatively long time to train (a few hours in the case here), they still offer many advantages over using a FOM. First, as discussed in the Introduction, it is often the case (e.g.~flows of complex fluids) that there is no exact FOM, where the point about how fast the FOM can be solved is not relevant; in this case, our hybrid ROMs can still be trained with timeseries data supplemented by approximate knowledge of the physics (we demonstrate this case later by using mismatched parameters between data and FOM). Second, the initial training time for these ROMs is easily offset once many evaluations of the ROM (or calculations of its Jacobians or adjoints) is performed\AGrevise{ --- this eventual return on investment is the main selling point of ROMs being applied to multi-query tasks.} Lastly, we comment that for the 1D PDE systems we consider in this paper (which we are just using as benchmarks), the ROM is not necessarily faster than the FOM, since the amount of dimension reduction from the ambient space dimension to the invariant manifold dimension is moderate --- by contrast, for systems such as a chaotic 2D or 3D flow where there is a vast amount of dimension reduction (e.g.~ambient space dimension of about 100,000 to manifold dimension 18), the speedup of a ROM would be more noticeable \cite{Linot.2023,Perez.2025.10.1103/ts3k-flx6,Jesús.2023.10.1103/physrevfluids.8.044402,Constante-Amores2025}.}

In Appendix \ref{sec:kse_lts}, we evaluate how well the solution trajectories stay on the true attractor at long times by calculating the joint probability density function (pdf) of some key quantities. There, we show that all five methods captured this long-time statistic equally well.

\subsection{Scarce data, exact FOM}
\label{sec:scarce_data_kse}

While still using the exact FOM, we now study the effect of having less data (training data size $N<400,000$), data spaced further apart ($\tau>0.25)$, or both. We apply this data-scarce training for the autoencoder, the MG-N network, and all NODEs. To prevent overfitting, the weight decay strength in all neural network training is increased here (see training details in Section \ref{sec:train_details}).

\begin{figure*}
	\centering
	\includegraphics[width=1\linewidth]{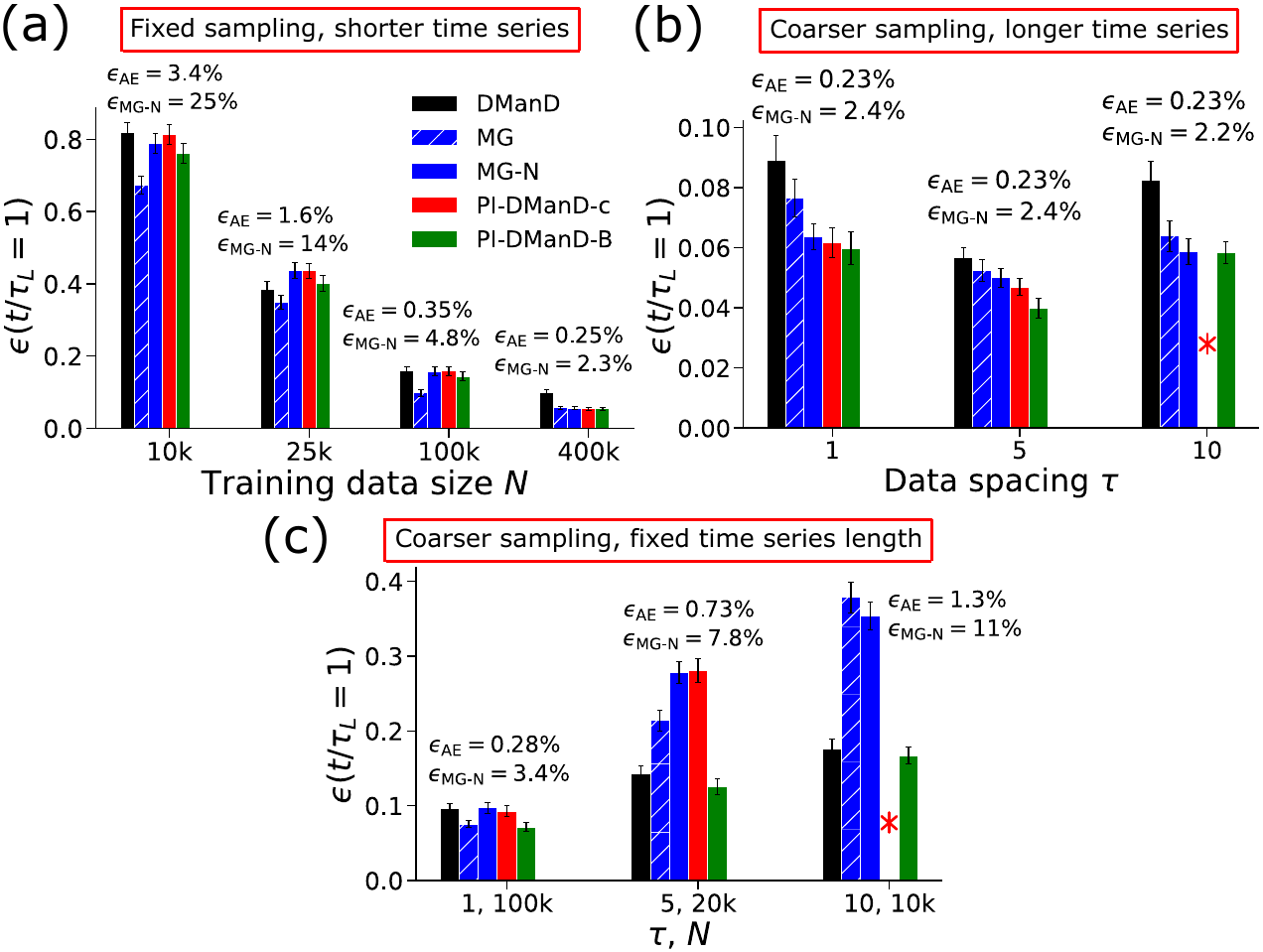}
    \caption{KSE, $L=22$ (for both data and FOM). Short-time tracking error at one Lyapunov time for scarce data in terms of (a) less data with $N \leq 400,000$ and $\tau=0.25$ (i.e.~shorter time series), (b) widely spaced data with $\tau > 0.25$ and $N=400,000$, and (c) increasing $\tau$ and decreasing $N$ proportionally. Here and elsewhere, error bars show standard error, and all subplots follow the legend in (a). Note the different vertical axis limits for each subplot. Missing PI-DManD-c bars, denoted with asterisk, in (b) and (c) have errors of about 1 (i.e.~much larger than the case shown) and are omitted to improve clarity in the vertical axis.}
	\label{fig:kse_vary}
\end{figure*}

We first fix the data spacing at $\tau=0.25$ and vary the training data size from $N=400,000$ to $100,000$, $25,000$, and $10,000$. In other words, we use a shorter part of our time series data. Figure \ref{fig:kse_vary}(a) shows the ensemble-averaged tracking error $\epsilon$ at one Lyapunov time for these four values of $N$. 
Above each group of bars in Figure \ref{fig:kse_vary}(a), we report the percent error for the autoencoder (AE) reconstruction
\begin{equation}
    \epsilon_\mathrm{AE} = \left\langle \frac{\| u_i - \tilde u_i \|_2}{\| u_i \|_2} \right\rangle_i \times 100 \% ,
    \label{eq:epsilon_ae}
\end{equation}
and the percent error for MG-N in approximating the MG vector field
\begin{equation}
    \epsilon_\MGN = \left\langle \frac{\| g_m(h_i) - g_{mn}(h_i) \|_2}{\| g_m(h_i) \|_2} \right\rangle_i \times 100 \% ,
    \label{eq:epsilon_mgn}
\end{equation}
where $\langle \cdot  \rangle_i$ is an ensemble average across all snapshots (indexed by $i$) in the testing dataset. The base case of $N=400,000$, which corresponds to Figure \ref{fig:kse_res}(a), shows that DManD has higher tracking error, by a factor of two, than the other methods.
Within the error bars, MG-N and the two hybrid methods perform comparably to DManD for $N < 400,000$. The reason that NODE test error $\epsilon$ increases with decreasing $N$ for all methods is due to increasing autoencoder reconstruction error.
\MDGrevise{ Also, MG performs somewhat better than the other methods for $N\leq 100,000$, presumably because autoencoder error is sufficiently small that it does not significantly affect the vector field on the manifold, and has smaller the vector-field approximation errors than MG-N and the DManD methods. We comment further on these issues below.}

Next, we fix $N=400,000$ and increase $\tau$ from 0.25 to 1, 5, and 10. As mentioned earlier, here we need to generate new time series data that cover a longer time interval than the originally used time series data of $N=400,000$ and $\tau=0.25$. As seen in Figure \ref{fig:kse_vary}(b), the autoencoder, MG-N, and NODE tracking errors do not vary much with $\tau$. Across the three increased $\tau$ values, PI-DManD-B has up to $30\%$ lower error than DManD. Note that the missing PI-DManD-c bar in Figure \ref{fig:kse_vary}(b), replaced by an asterisk, has an error of about unity and is omitted to improve clarity in the vertical axis (and similarly for bar plots elsewhere). Note that a spacing of $\tau=10$ corresponds to $\approx0.5$ Lyapunov times, and all approaches will ultimately fail as $\tau$ becomes comparable to the Lyapunov time; we find that PI-DManD-c fails first, possibly since it must learn an additive correction to the fixed MG-N vector field, while PI-DManD-B dynamically updates the MG-N weights with data. Unlike the previously considered scenario with just decreasing $N$, here we see sizable improvements of PI-DManD-B over DManD.

In the final subcase here, \MDGrevise{we downsample the original time series dataset ($N=400,000$ and $\tau=0.25$), holding the time series length constant, but increasing $\tau$ so that the total number $N$ of data vectors decreases.}. In Figure \ref{fig:kse_vary}(c), we show the tracking error for downsample factors of $4$, $20$, and $40$; here, only PI-DManD-B has a tracking error that remains comparable to that of DManD (and is slightly lower in the case of $\tau=1$, $N=100,000$). \MDGrevise{As the data becomes more sparse, the autoencoder error increases substantially. Since the MG and MG-N methods do not access the time-series nature of the data, they rely on evaluations of the FOM vector field at positions that, because of autoencoder error, do not lie on the manifold. By contrast, DManD and PI-DManD learn the vector field on the approximate manifold that best reconstructs the time series data in the ambient space, so are less sensitive to autoencoder error. Relatedly, in past work, we have shown that for the KSE, highly accurate DManD models can be found even if the encoder is a linear projection \cite{Linot.2022.10.1063/5.0069536}. Linear encoders are explored more generally in Ref.~\onlinecite{Pottier20261048550arxiv260829867zc}. In general, what is required for a linear encoder to work is that there exists an injective linear projection of the manifold onto the latent state -- i.e. that a point on the manifold can be uniquely mapped to a linear subspace. This observation implies that it is not necessary to impose a tangency condition on the autoencoder to obtain good time series prediction performance.}

\MDGrevise{In summary, across these three subcases of data-scarce training, PI-DManD-B outperforms MG-N and has equal or lower error than DManD. We now use PI-DManD to incorporate data-driven information when the physics-based vector field is imperfect.}


\subsection{Approximate FOM}
\label{sec:kse_err_FOM}

\begin{figure*}
	\centering
	\includegraphics[width=1\linewidth]{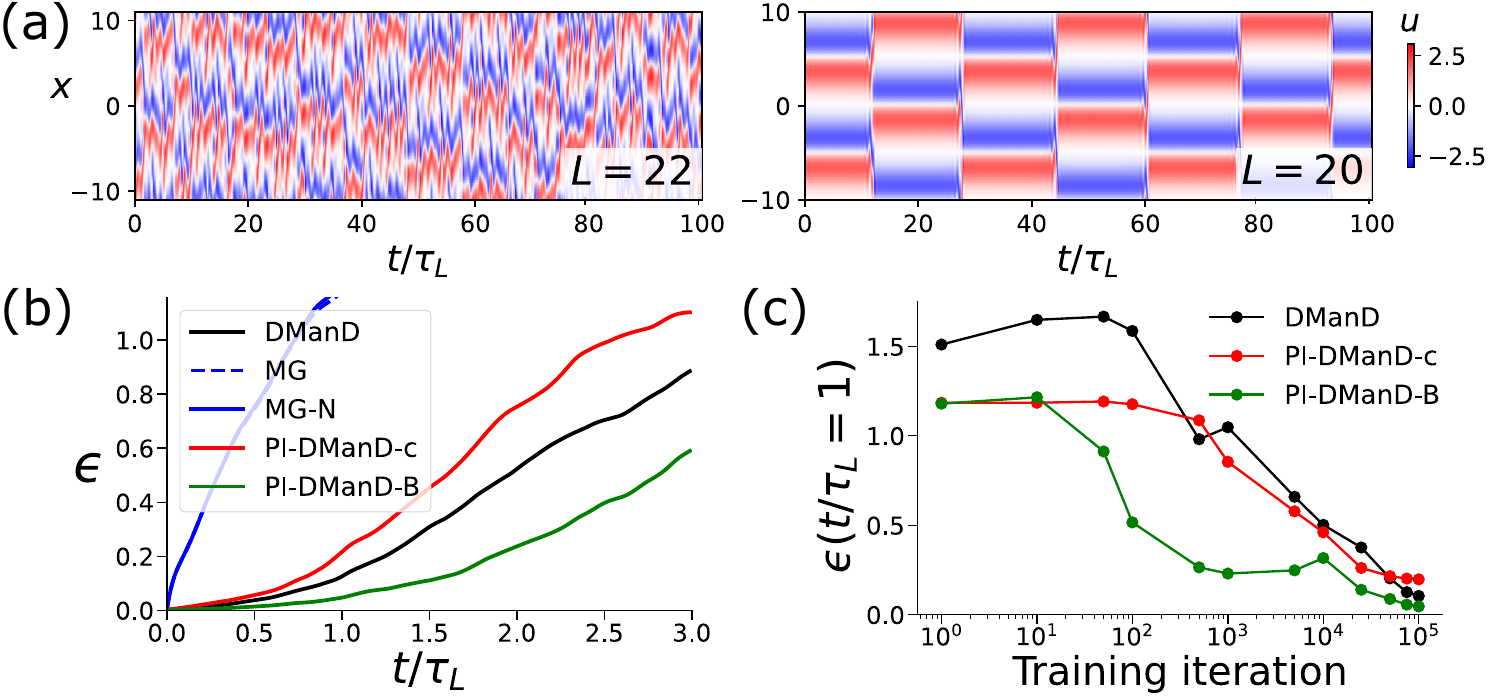}
	\caption{KSE, $N=400,000$, $\tau=0.25$. Effects of using an erroneous parameter ($L=20$) in the FOM to model data generated with $L=22$. (a) Ground truth trajectory (left) and example trajectory with erroneous parameter (right). (b) Ensemble-averaged short-time tracking error. (c) NODE training convergence for purely data-driven and hybrid methods.}
	\label{fig:kse_L20}
\end{figure*}

An exact FOM is not available for many physical systems, either because of a lack of first-principles knowledge of the governing equations, or uncertainty in the parameter values used in the FOM. Here we test the robustness of our hybrid methods to model error by purposely using an incorrect domain length $L = 20$ in the FOM instead of the correct value of $L=22$ used to generate the data (i.e.~$\mu_\FOM \neq \mu$). In other words, the model manifold vector field $g_m$ is generated from $f_\FOM$ with incorrect parameter values. Figure \ref{fig:kse_L20}(a) shows the space-time plots of example trajectories for $L=22$ and $L=20$. With the smaller domain length, the dynamics display intermittent/bursting-like behavior that is very different from that found at $L=22$. To be clear, we never generate data with $L=20$ for any training or testing of our models --- we only use the $L=22$ dataset.
\AG{The parameter $L=20$ comes into play only when we feed the $L=22$ dataset into the $L=20$ FOM.}
\AGrevise{Specifically, we apply the KSE operator with $L=20$ to the Fourier-transformed state snapshots generated with $L=22$. The state snapshots are unchanged, but the operator used to evaluate their spatial derivatives --- and consequently the evaluated vector field --- has different wavenumbers.}

To make the comparisons in this section, we use abundant training data ($N=400,000$ and $\tau=0.25$). Figure \ref{fig:kse_L20}(b) shows the tracking error for all methods. Since DManD does not use any knowledge of the FOM, its short-time tracking performance is comparable to that in Figure \ref{fig:kse_res}(a) (these two NODE models for DManD only differ by the random weight initialization). As expected, the solely physics-informed MG and MG-N methods have poor short-time tracking due to only trying to learn the vector field in the latent coordinates from the FOM (see Equation \ref{eq:MG}) and not from the data. 
Since PI-DManD-c uses data to correct for FOM errors, its tracking error is lower than that of MG-N; as shown in Figure \ref{fig:kse_L20}(c), the NODE training convergence for $g_c$ of PI-DManD-c converges slower as compared to the exact FOM case in Figure \ref{fig:kse_res}, indicating that correcting FOM errors requires more training iterations than only correcting for numerical errors. While PI-DManD-c does not outperform DManD in short-time tracking, PI-DManD-B does. This result indicates that even if the FOM of the physics-based prior is highly erroneous, the hybrid PI-DManD-B ROM can still substantially outperform the solely data-driven DManD ROM in short-time tracking.
\AG{The mechanism for this result is essentially \AGrevise{from the warm start}, where the initialization with a physics-based vector field (even if it is from an approximate FOM) is much better than a random guess when training a NODE.}
As for long-time statistics, both hybrid ROMs perform similarly to DManD (see Appendix \ref{sec:kse_lts}).

\section{Conclusion} 
\label{sec:conc}

In this work, we found that including physics-based prior information substantially improves the predictive capabilities of data-driven ROMs applied to systems with spatiotemporal chaotic dynamics, even if the physics-based information is not completely correct. We built upon previous work on a solely data-driven ROM called DManD \cite{Linot2020} --- which uses a neural ODE (NODE) approach to learn the dynamics on an invariant manifold whose coordinates are found with an autoencoder --- by incorporating ideas from the Manifold Galerkin (MG) method \cite{Lee2020}. MG expresses the RHS of the ODE in the latent space coordinates in terms of the RHS of the FOM (i.e.~the physics).
We approximated the MG vector field with a neural network (which we call MG-N) and combined it with DManD to develop two hybrid ROMs.
The first, which we call PI-DManD-c, adds a correction term learned by a NODE to the MG-N vector field in order to account for numerical errors and inherent errors in the FOM. The second hybrid ROM is a DManD model with an additional weight-decay-like term in the loss that drives the weights of a NODE network toward those of the MG-N network (and \AGrevise{warm-starting the former with weights} of the latter). This second hybrid ROM can be viewed as using MG-N as a Bayesian prior for training a NODE, and we call it PI-DManD-B (note, while the method is motivated using Bayes' theorem, it is not probabilistic).

We applied DManD and our hybrid ROMs (PI-DManD-c and PI-DManD-B) to the KSE and the 1D CGLE (results for CGLE are in Appendix \ref{sec:cgle}), which are two PDE systems with spatiotemporal chaotic dynamics. We summarize the improvement of PI-DManD-B over DManD in making short-time predictions for three case studies in Table \ref{table:conc}.
The first case study is the baseline, where we use exact physics and abundant training data; there, PI-DManD-B (and PI-DManD-c) accurately tracked the true solution for substantially longer --- at one Lyapunov time, the ensemble-averaged relative short-time tracking error is $0.1$ for DManD and only $0.05$ for the hybrid ROMs. The second case study reduced the amount of training data (for three subcases), and the third study involved providing substantially incorrect PDE parameters to the FOM used by MG/MG-N. For these last two studies, PI-DManD-B (but not PI-DManD-C) still provided sizable improvements over DManD, thus demonstrating its robustness to scarce data and FOM errors.
\AG{While the total training time (MG-N and NODE) for PI-DManD-B is somewhat longer than PI-DManD-c, we recommend PI-DManD-B due to its better short-time tracking performance.}
Future directions include testing robustness of our hybrid ROMs on complex fluid and flow systems for which only approximate first-principles models are available, and extending our methods to the case where we only have partial observations. \MDGrevise{Another avenue of future work is the application of PI-DManD to systems with much larger ambient and manifold dimensions. An unsolved challenge here is finding highly-accurate autoencoders, as these underlie the method.}
\AG{Lastly, in other future work, we aim to apply our methods to the real-world scenario of noisy data, where we expect the physics-based information to aid in evolving the dynamics on the underlying invariant manifold.}


\begin{table*}
    \setlength{\tabcolsep}{2pt}
    \centering
    \begin{tabular}{c|ccc}
    \phantom & \specialcell{Exact physics,\\ abundant data \\[3pt]} & \specialcell{Exact physics,\\ scarce data \\[3pt]} & \specialcell{Erroneous physics,\\ abundant data \\[3pt]} \\
    \hline
    \specialcell{\% reduction \\ in relative \\ trajectory \\ error at \\ $t=\tau_L$} & \
    \specialcell{\textbf{46\%--51\%} \\ (Figs. \ref{fig:kse_res}(a), \\ \ref{fig:cgle_res}(a))} &\
    \specialcell{\\[-8pt] \textbf{None} (shorter time series, finer \\ sampling; Figs. \ref{fig:kse_vary}(a), \ref{fig:cgle_vary}(a)) \\[5pt] \textbf{22\%--62\%} (longer time series, \\ coarser sampling; Figs. \ref{fig:kse_vary}(b), \ref{fig:cgle_vary}(b)) \\[5pt] \textbf{6\%--29\%} (fixed time series length, \\ coarser sampling; Figs. \ref{fig:kse_vary}(c), \ref{fig:cgle_vary}(c))} &\
    \specialcell{\textbf{43\%--64\%} \\ (Figs. \ref{fig:kse_L20}(b), \\ \ref{fig:cgle_vary_beta}(b), \ref{fig:cgle_vary_beta}(c))}
    \end{tabular}
    \caption{Summary of the improvement or reduction in $\epsilon(t=\tau_L)$ (the relative ensemble averaged short-time tracking error at one Lyapunov time) for the hybrid PI-DManD-B method over the solely data-driven DManD method; specifically, the quantities correspond to $(\epsilon_\text{DManD} - \epsilon_\text{PI-DManD-B}) / \epsilon_\text{DManD} \times 100\%$. We varied the accuracy of the known physics (i.e.~the full order model), the length of the time series (i.e.~number of time units covered), and the sampling rate of the data.}
    \label{table:conc}
\end{table*}

\section*{Acknowledgments}
{We thank Jake Buzhardt for helpful discussions. This work was supported by the National Science Foundation (CBET-2347344), Office of Naval Research (N00014-18-1-2865 (Vannevar Bush Faculty Fellowship)), and Air Force Office of Scientific Research (FA9550-26-1-B007).}

\section*{Author Declarations}
\subsection*{Conflict of Interest}
The authors have no conflicts to disclose.

\section*{Author contributions}
M.D.G. obtained funding and conceptualized the research. A.G. performed the research and wrote the main manuscript text. Both authors reviewed the manuscript.

\section*{Data availability statement}
The data that support the findings of this study are available from the corresponding author upon reasonable request.

\appendix
\section*{Appendix}

\renewcommand{\thesubsection}{\Alph{subsection}}
\renewcommand{\thesubsubsection}{\Roman{subsubsection}}




\subsection{Training and testing speed for methods applied to KSE}
\label{sec:kse_train_speed}

Here, we examine the cost of training and evaluating each method. As shown in Figure \ref{fig:kse_res}(d), the NODE $g_c$ learned by the PI-DManD-c model trains orders of magnitude faster than the DManD model --- PI-DManD-c requires only about 10 training iterations for convergence, while DManD requires $>10^4$ iterations (PI-DManD-B also takes a similar number of iterations as DManD). However, the PI-DManD-c model adds $g_c$ to the MG-N model $g_{mn}$ (see Equation \ref{eq:pidmand}), so the time to obtain an MG-N model needs to be considered in addition to the NODE training time. We define training time for the hybrid ROMs as the time needed to obtain a trained MG-N model plus the time needed to train the NODE. We also define training time for MG-N as the network training time plus the time for preprocessing the MG vector field output data from the data of the vector field on the state space (0.1 hours for KSE). Table \ref{table:speed} shows the training time for all methods.
One can see that the total time for training either hybrid methods is greater than that of DManD; however, as seen in the last row of Table \ref{table:speed}, MG-N and the hybrid methods are comparable to DManD in regards to how fast a test trajectory can be evaluated for 100 Lyapunov times. While MG requires no training time to obtain the vector field on the manifold, it is over an order of magnitude slower to evaluate than the other methods --- as discussed in Section \ref{sec:mg}, the reason is that MG needs to decode $h$, solve the FOM, and evaluate a Jacobian (which involves an encoding step) for each function evaluation of $g_m$. So, MG is clearly not a ROM. Although the MG-N and hybrid methods require a longer time to train than DManD, they all take a similar amount of time to evaluate/test
and have the added benefit of substantially better short-time tracking of the true solution.

\begin{table*}[t]
\renewcommand{\arraystretch}{1.5}
\centering
\setlength{\tabcolsep}{10pt}
\begin{tabular}{ c|ccccc } 
     \phantom & DManD & MG-N & MG & PI-DManD-c & PI-DManD-B \\
     \hline
     Train time (hr) & 2.2 & 8.3 & n/a & 13.1 & 12.0 \\
     Test time (s) & 8.3 & 8.2 & 102 & 11.4 & 7.7
\end{tabular}
\caption{Comparing computation time for methods applied to KSE. First row is time (in hours) for training. Second row is computation time (in seconds) for testing, where an initial condition is integrated for 100 Lyapunov times. Calculations were run on an Intel Xeon E5-2640 CPU.}
\label{table:speed}
\end{table*}

\subsection{Long time statistics results for KSE}
\label{sec:kse_lts}

\begin{figure*}
	\centering
	\includegraphics[width=1\linewidth]{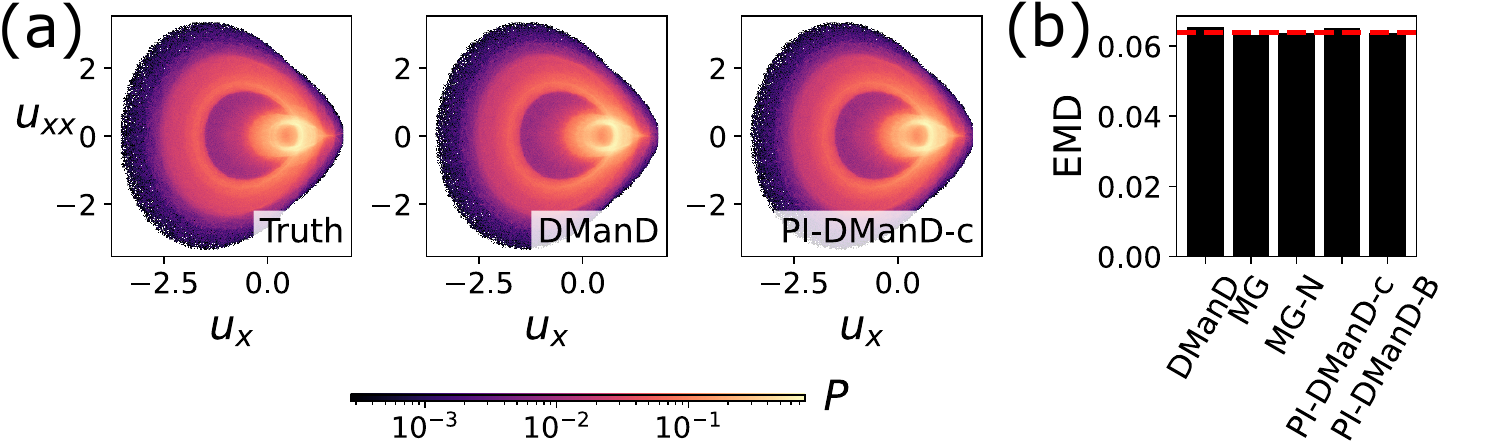}
	\caption{KSE, $L=22$ (for both data and FOM), $N=400,000$, $\tau=0.25$. (a) Comparison of joint pdfs. (b) Earth movers distance (EMD) quantifying the difference between pdfs of the ground truth and model predictions, where the dashed red line is the baseline EMD calculated with the pdf of another ground truth trajectory. Joint pdfs for models not shown look similar to joint pdfs of models with similar EMD values.}
	\label{fig:kse_lts}
\end{figure*}

Besides the short-time tracking results described in Section \ref{sec:kse_lots_data}, we also gauge model performance based on how well a predicted trajectory stays on the true attractor at long times. Since for KSE at $L=22$ the invariant manifold for the long-time dynamics is 8-dimensional, we simplify the comparison of the long-time statistics by choosing two relevant quantities and calculating a joint probability distribution function (pdf). The time-evolution of the ``energy'' $\overline{u^2}/2$, where $\overline \cdot$ denotes  average over the spatial domain, has a growth (``production") term proportional to $||\overline{u_x^2}||_2$ and a decay (``dissipation") term proportional to $\overline{u_{xx}^2}$, where the subscript $x$ denotes spatial derivative. Based on this observation, we use the joint pdf of $u_x$ and $u_{xx}$ to characterize the attractor. We calculate joint pdfs using the entire testing dataset (100,000 snapshots spaced by $\tau=0.25$ time units), where model predictions use the first snapshot as an initial condition. Figure \ref{fig:kse_lts}(a) shows the joint pdfs for long-time trajectories corresponding to the ground truth, DManD, and PI-DManD-c. The joint pdfs for PI-DManD-B and the two physics-informed methods are not shown since they are virtually identical to the joint pdfs that are shown. To quantitatively compare the pdfs of the model predictions to the pdf of the ground truth trajectory, we calculate the earth mover's distance (EMD), also known as the Wasserstein-1 distance \cite{Peyr2018}. Let $\mu$ and $\nu$ be two discrete distributions calculated from normalized histograms with $n_\mu$ and $n_\nu$ bins, respectively. The EMD between these two distributions is the minimal cost of transforming one distribution to the other, and is given by
\begin{align}
    \mathrm{EMD}(\mu,\nu) = \min_\gamma \sum_{i,j} \gamma_{i,j} \left \| \mu^{(i)} - \nu^{(j)} \right \|_2 \\
    \mathrm{s.t.} \quad \gamma \textbf{1} = \mu , \quad \gamma^\top \textbf{1} = \nu , \quad \gamma \geq 0 ,
    \label{eq:emd}
\end{align}
where $\textbf{1}$ is a column vector of ones and $\gamma \in \mathbb R^{n_\mu \times n_\nu}$ is a transport plan with $\gamma_{i,j}$ dictating the amount of probability mass to move from bin $\mu^{(i)}$ to bin $\nu^{(j)}$. The EMD represents the minimal work (amount of mass moved multiplied by distance) needed to execute the optimal transport plan. Throughout this paper, the EMD value is calculated with respect to the pdf of the ground truth trajectory in the testing dataset, which we call $\mu_t$.

We first show the long-time statistics reconstruction for the case of abundant data and exact FOM (see Section \ref{sec:kse_lots_data}). In Figure \ref{fig:kse_lts}(b), we plot the EMD between $\mu_t$ and the model prediction pdf for all five methods. Note that the dashed red line in Figure \ref{fig:kse_lts}(b) represents the baseline EMD, which is calculated between $\mu_t$ and the pdf of a ground truth trajectory in the training dataset. Since all methods have a EMD value similar to this baseline EMD, they perform equally well in capturing the long-time statistics, at least for the joint pdf quantities chosen here.

\begin{figure*}
	\centering
	\includegraphics[width=1\linewidth]{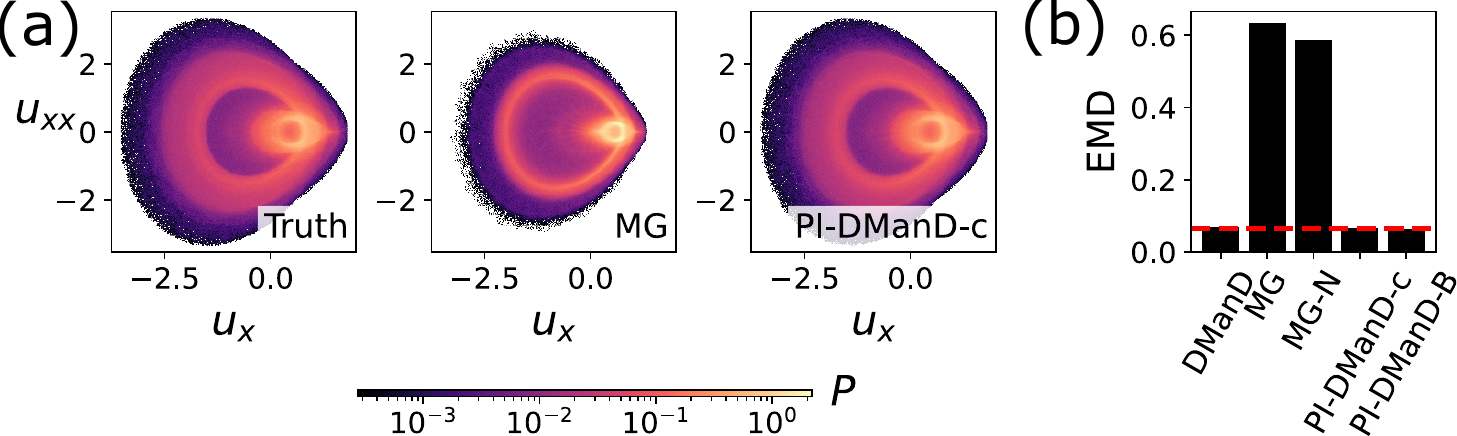}
	\caption{KSE, $N=400,000$, $\tau=0.25$. Effects of using an erroneous parameter ($L=20$) in the FOM when making model predictions on the data generated with $L=22$. (a) Comparison of joint pdfs. (b) Earth movers distance (EMD) quantifying the difference between pdfs of the ground truth and model predictions, where the dashed red line is the baseline EMD calculated with the pdf of another ground truth trajectory. Joint pdfs for models not shown look similar to joint pdfs of models with similar EMD values.}
	\label{fig:kse_L20_lts}
\end{figure*}

Next, we show the long-time statistics reconstruction for the case of abundant data and inexact FOM (see Section \ref{sec:kse_err_FOM}). In Figure \ref{fig:kse_L20_lts}(a), we plot the joint pdfs for the ground truth trajectory and the MG and PI-DManD-c model predictions. In Figure \ref{fig:kse_L20_lts}(b), we plot the EMD value (see Equation \ref{eq:emd}) between pdfs of a model prediction and the ground truth test trajectory; again, the baseline EMD (dashed red line in the figure) is calculated between two ground truth trajectories. (Pdfs for models not shown are similar to shown pdfs if they have similar EMD values.) Figure \ref{fig:kse_L20_lts} shows that when given an erroneous parameter value for $f_\FOM$, MG and MG-N will evolve the $L=22$ dataset on a completely incorrect attractor. As for the two hybrid methods, they capture the long-time statistics as well as DManD.

\subsection{Complex Ginzburg--Landau Equation}
\label{sec:cgle}

In addition to the KSE, we have studied PI-DManD for prediction of dynamics of the one-dimensional complex Ginzburg--Landau equation (CGLE)
\begin{equation}
    \frac{\partial A}{\partial t} = A + (1 + i \alpha)\frac{\partial^2 A}{\partial x^2} - (1+ i \beta) |A|^2 A \, .
    \label{eq:cgle}
\end{equation}
\AG{The CGLE has been used to model reaction-diffusion systems, superfluidity, and liquid crystals \cite{Aranson2002}.}
The quantities $\alpha$ and $\beta$ are parameters. We use a domain length $L=30$ and periodic boundary conditions. We set $\alpha=2$ and $\beta=-2$, so when $L=30$ the dynamics lie in the ``defect-mediated turbulence" regime \cite{Shraiman1992}. The solution is computed with an ETD-RK4 solver \cite{Kassam2005} in the Fourier domain. We used the Bennitin algorithm \cite{Benettin1980} to determine a Lyapunov time of $2.8$ time units.
With the initial transient removed, we generate $10^5$ snapshots of $A \in \mathbb C^{64}$ discretized on $d_u=64$ uniformly spaced grid points and separated by $\tau=0.05$ time units (note, $\tau$ is five times smaller here for the CGLE compared to the KSE). To work with real numbers in our neural networks\AGrevise{, for simplicity}, we concatenate the real and imaginary parts of $A$ to form $u \in \mathbb R^{128}$. We use $u$ when calculating error quantities.

\subsubsection{Determining manifold dimension}
\label{sec:cgle_dm}

Unlike the KSE system considered previously, the dimension of the invariant manifold $d_\mathcal M$ for data on the attractor for this CGLE system is not known. To estimate it, we use an implicit rank-minimizing autoencoder with weight decay (IRMAE-WD) \cite{jing2020implicit,irmae-c}. IRMAE-WD uses an intermediate latent space $z \in \mathbb R^{d_z}$, where $d_z$ acts as an upper bound on the manifold dimension (we use $d_z=50$ here). The difference between IRMAE-WD and a normal autoencoder is the use of linear layers in the former, which act to drive the covariance matrix $zz^T$ to have minimal rank $d_\mathcal M < d_z$. IRMAE-WD has been applied to the KSE in Ref.~\onlinecite{irmae-c}, where the singular value spectra of $zz^T$ contains a sharp drop-off that spans over 10 orders of magnitude --- this clear indication of the rank of $zz^T$ then gives us an estimate of the manifold dimension $d_\mathcal M$.

\begin{figure}
	\centering
	\includegraphics[width=0.8\linewidth]{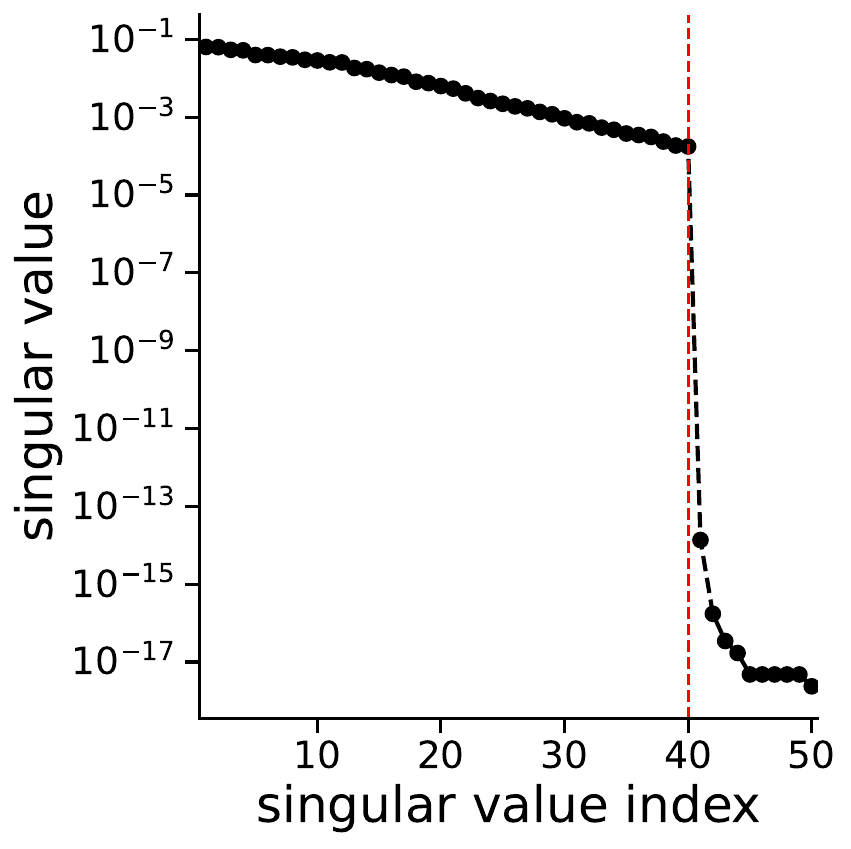}
	\caption{Singular value spectra of the covariance of the latent space $z$ for IRMAE-WD applied to CGLE. Red dashed line located at 40th singular value index.}
	\label{fig:irmae_cgle}
\end{figure}

We train an IRMAE-WD autoencoder on the CGLE system with weight-decay strength $10^{-5}$ and four linear layers. Figure \ref{fig:irmae_cgle} shows the singular value spectra, where the sharp drop-off covering about 10 orders of magnitude occurring at the 40th singular value index indicates that our data has $d_\mathcal M = 40$. The IRMAE-WD relative reconstruction error (calculated with Equation \ref{eq:epsilon_ae}) is $\epsilon_\mathrm{AE}=0.61\%$. Note that we observed similarly sharp drop-offs at singular value indices ranging from 33 to 39 for 4 other IRMAE models. Therefore, we choose an estimate that is at the high end (40) of an ensemble of estimates.

Before using the IRMAE-WD in a ROM, we would need to project the intermediate latent space $z \in \mathbb R^{50}$ onto a latent space $h \in \mathbb R^{40}$ (e.g.~with a singular value decomposition); however, doing so would make the encoder $\Enc: u \rightarrow h$ more complicated. Therefore, to simplify the encoder structure whose Jacobian $J_\Enc(h)$ must be computed for MG, we choose to instead train and use a normal autoencoder with latent dimension $d_h = 40$. The relative reconstruction error for this normal autoencoder is $\epsilon_\mathrm{AE}=0.57\%$, which is similar to that of the IRMAE-WD autoencoder.
All of the training details --- such as neural network architectures and learning rate scheduling --- are identical to those for the KSE, which are shown in Table \ref{table:arch} and described in the body of Section \ref{sec:train_details}.

\subsubsection{Abundant data, exact FOM}
\label{sec:cgle_lots_data}

\begin{figure*}
	\centering
	\includegraphics[width=1\linewidth]{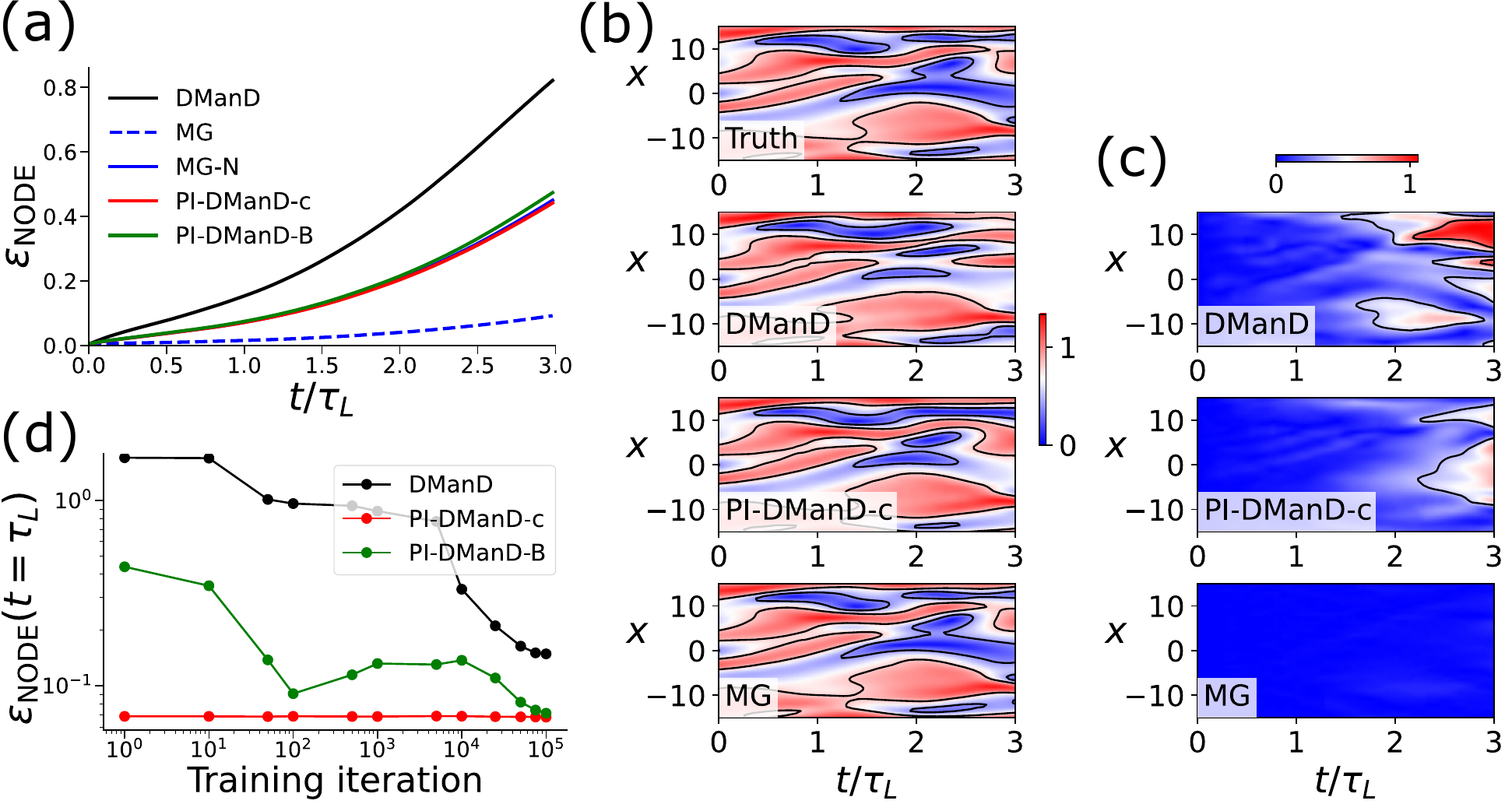}
	\caption{CGLE, $\beta=-2$ (for both data and FOM), $N=400,000$, $\tau=0.05$. (a) Ensemble-averaged short-time tracking error for all methods. (b) Absolute value of representative trajectories of ground truth and model predictions (contours at $|A| = 0.4,0.8$). (c) Difference plots between ground truth and model predictions (contours at difference values of 0.4, 0.8). Representative trajectories and difference plots for other models look similar to those of PI-DManD-c. (d) NODE training convergence for purely data-driven and hybrid methods (note the log-log scale).}
	\label{fig:cgle_res}
\end{figure*}

We begin presentation of the results by paralleling the KSE results of Section \ref{sec:kse_lots_data}, using the entire dataset of $N=400,000$ training snapshots spaced apart by $\tau=0.05$ time units and using the same parameters $\alpha=2$ and $\beta=-2$ for data generation and the FOM.
Figure \ref{fig:cgle_res}(a) shows the ensemble-averaged tracking error for all methods. The physics-informed MG-N method (whose curve is directly underneath the red PI-DManD-c curve in Figure \ref{fig:cgle_res}(a)) and the two hybrid methods outperform DManD in short-time tracking; while this result is consistent with the KSE results in Figure \ref{fig:kse_res}(a), the primary difference here is that MG has dramatically lower short-time tracking error. As shown in the representative space-time plots in Figure \ref{fig:cgle_res}(b) and the difference plots in Figure \ref{fig:cgle_res}(c), MG is tracking the true solution with very small error for 3 Lyapunov times.
For KSE, the short-time tracking error at one Lyapunov time is about 0.07 for both MG and MG-N; however, for CGLE, it is about 0.07 for MG-N and about 0.02 for MG.
We believe that the gap in the error between MG and MG-N is larger for the CGLE case due to a higher error in approximating $g_m$ with $g_{mn}$ ($\epsilon_\MGN$ is $8\%$ for CGLE and $2.3\%$ for KSE).



The superior predictive performance of MG for the CGLE comes at a large cost.
In Table \ref{table:speed_cgle}, we show the training and testing time for all methods (analagous to Table \ref{table:speed} for the KSE). Here, MG still takes almost two orders of magnitude longer to evolve a trajectory.
Figure \ref{fig:cgle_res}(d) shows the NODE training convergence, where (like with the KSE) the NODE $g_c$ learned by the PI-DManD-c model training orders of magnitude faster than the DManD model. However, when accounting for the training cost of the MG-N network $g_{mn}$ (which includes 0.15 hours of preprocessing time), the total cost for training both $g_c$ and $g_{mn}$ for PI-DManD-c is greater than for DManD (see Table \ref{table:speed_cgle}).

\begin{table*}
\renewcommand{\arraystretch}{1.5}
\centering
\setlength{\tabcolsep}{10pt}
\begin{tabular}{ c|ccccc } 
     \phantom & DManD & MG-N & MG & PI-DManD-c & PI-DManD-B \\
     \hline
     Train time (hr) & 2.2 & 9.3 & n/a & 13.0 & 12.3 \\
     Test time (s) & 4.9 & 5.0 & 218.0 & 6.8 & 4.5
\end{tabular}
\caption{Comparing computation time for methods applied to CGLE. First row is time (in hours) for training. Second row is computation time (in seconds) for testing, where an initial condition is integrated for 100 Lyapunov times. Calculations were run on an Intel Xeon E5-2640 CPU.}
\label{table:speed_cgle}
\end{table*}

\begin{figure*}
	\centering
	\includegraphics[width=1\linewidth]{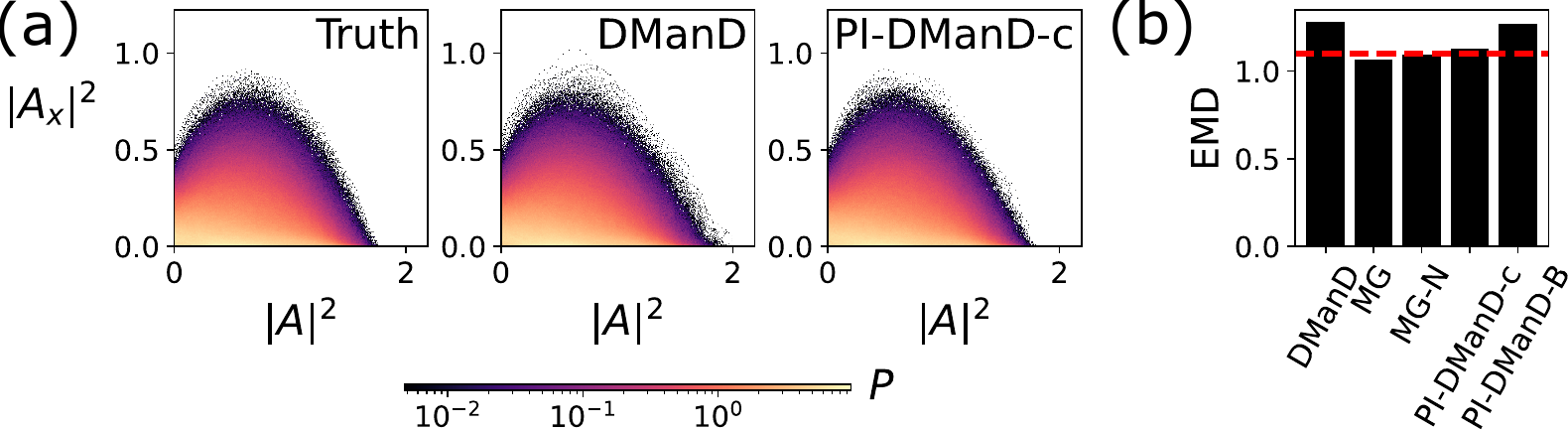}
	\caption{CGLE, $\beta=-2$ (for both data and FOM), $N=400,000$, $\tau=0.05$. (a) Comparison of joint pdfs. (b) Earth movers distance (EMD) quantifying the difference between pdfs of the ground truth and model predictions, where the dashed red line is the baseline EMD calculated with the pdf of another ground truth trajectory. Joint pdfs for models not shown look similar to joint pdfs of models with similar EMD values.}
	\label{fig:cgle_lts}
\end{figure*}

We now determine whether long-time trajectories of each method stay on the true CGLE attractor. As with the KSE, we will calculate a joint pdf to use as a simple metric for the long-time statistics. We again performed a
kinetic energy ($|A|^2$) balance
on the PDE, which yielded a production term $\overline{|A|^2}$ and a dissipation term $\overline{|A_x|^2+|A|^4}$, where $A_x$ is a spatial derivative; from these terms, we choose the joint pdf quantities as $|A|^2$ and $|A_x|^2$, respectively. The joint pdfs for the ground truth, DManD and PI-DManD-c model predictions are shown in Figure \ref{fig:cgle_lts}(a), while the EMD (see Equation \ref{eq:emd}) calculated between all methods and the ground truth trajectory is presented in Figure \ref{fig:cgle_lts}(b). As before, the baseline EMD for the dashed red line is calculated between ground truth trajectories with different initial conditions. All methods have a similar EMD to this baseline, so they perform equally well in capturing the long-time statistics of these quantities.

\subsubsection{Scarce data, exact FOM}
\label{sec:scarce_data_cgle}

Here, we parallel the KSE results of Section \ref{sec:scarce_data_kse}. As with the KSE, we use the exact FOM and study the effect of having less data (training data size $N<400,000$), data spaced further apart ($\tau>0.05)$, or both simultaneously.

\begin{figure*}
	\centering
	\includegraphics[width=1\linewidth]{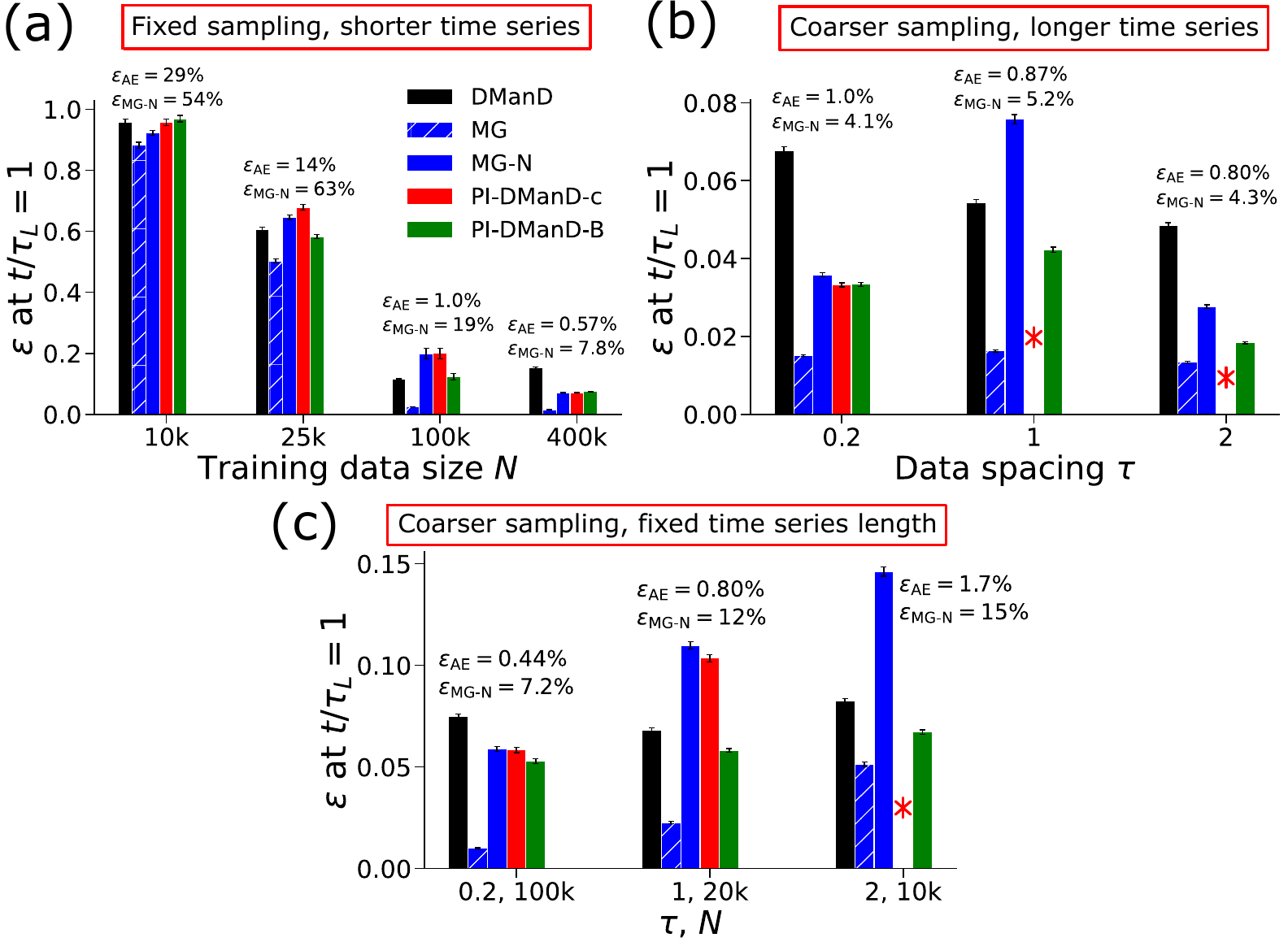}
	\caption{CGLE, $\beta=-2$ (for both data and FOM). Short-time tracking error at one Lyapunov time for scarce data in terms of (a) less data with $N \leq 400,000$ and $\tau=0.05$ (i.e.~shorter time series), (b) widely spaced data with $\tau > 0.05$ and $N=400,000$, and (c) increasing $\tau$ and decreasing $N$ proportionally. Error bars are standard errors, and all subplots follow the legend in (a). Note the different vertical axis limits for each subplot. Missing PI-DManD-c bars, denoted with asterisk, in (b) and (c) have errors of about 1 and are omitted to improve clarity in vertical axis.}
	\label{fig:cgle_vary}
\end{figure*}

We first fix the data spacing at $\tau=0.05$ and vary the training data size from $N=400,000$ to 100,000, 25,000, and 10,000 (i.e.~shorter time series). Figure \ref{fig:cgle_vary}(a) shows the ensemble-averaged tracking error $\epsilon$ at one Lyapunov time for decreasing $N$ (the case of $N=400,000$ corresponds to Figure \ref{fig:cgle_res}). Similar to the KSE, here the NODE test error $\epsilon$ increases with decreasing $N$ for all methods due to increasing autoencoder reconstruction error $\epsilon_\mathrm{AE}$. The autoencoder errors $\epsilon_\mathrm{AE}$ and MG-N errors $\epsilon_\MGN$ here are higher than they were for the KSE, but the NODE errors $\epsilon$ are about the same for both datasets. Another difference is that for $N=100,000$, PI-DManD-B (but not PI-DManD-c) is able to achieve error that is substantially lower than that of MG-N, which suggests the hybrid ROM is correcting for MG-N test error.

The second scenario is using a fixed number of training snapshots $N=400,000$ and increasing the data spacing from $\tau=0.05$ to 0.2, 1, and 2 (i.e.~use time series that are longer and sampled less often as $\tau$ increases). Figure \ref{fig:cgle_vary}(b) shows the errors for this scenario. As with the KSE, here the autoencoder, MG-N, and NODE errors do not vary much with $\tau$. Note that the missing PI-DManD-c bars in Figure \ref{fig:cgle_vary}(b), denoted with asterisk, have an error of about 1 and are omitted to improve clarity in the vertical axis. The reason is that $\tau=1$ time units corresponds to about one third of a Lyapunov time, so PI-DManD-c struggles in short-time tracking in this case and for $\tau=2$. Across the three increased $\tau$ values, both hybrid ROMs have lower error than DManD (except $\tau=1$ and 2, where only PI-DManD-B does).

Lastly, we study the effect of both increasing the data spacing $\tau$ and decreasing the number of training snapshots $N$ (i.e.~downsampling the original time series of length $N=400,000$ and $\tau=0.05$). In Figure \ref{fig:cgle_vary}(c), we show the tracking error for data spacing $\tau=0.2$, 1, and 2 with $N$ decreasing proportionally. The tracking error $\epsilon$ increases for all methods with increasing $\tau$ due to the increasing autoencoder error $\epsilon_\mathrm{AE}$. (As with the KSE, the PI-DManD-c error for $\tau=2$ and $N=10,000$ is around 1 and is omitted from Figure \ref{fig:cgle_vary}(c) to improve the clarity in the scaling of the vertical axis.) As with the KSE, the tracking errors for MG, MG-N, and PI-DManD-c become larger than that of DManD. For the KSE results in Figure \ref{fig:kse_vary}(c), the PI-DManD-B error remained comparable to DManD error for two out of the three values of $\tau$ and was slightly less than the DManD error for the other value of $\tau$; here, however, PI-DManD-B has somewhat lower error than DManD for all three values of $\tau$. As with the KSE, across the three cases of data-scarce training,
PI-DManD-B always has equal or lower error than DManD while PI-DManD-c sometimes has a higher error.

\subsubsection{Approximate FOM}

\begin{figure*}
	\centering
	\includegraphics[width=1\linewidth]{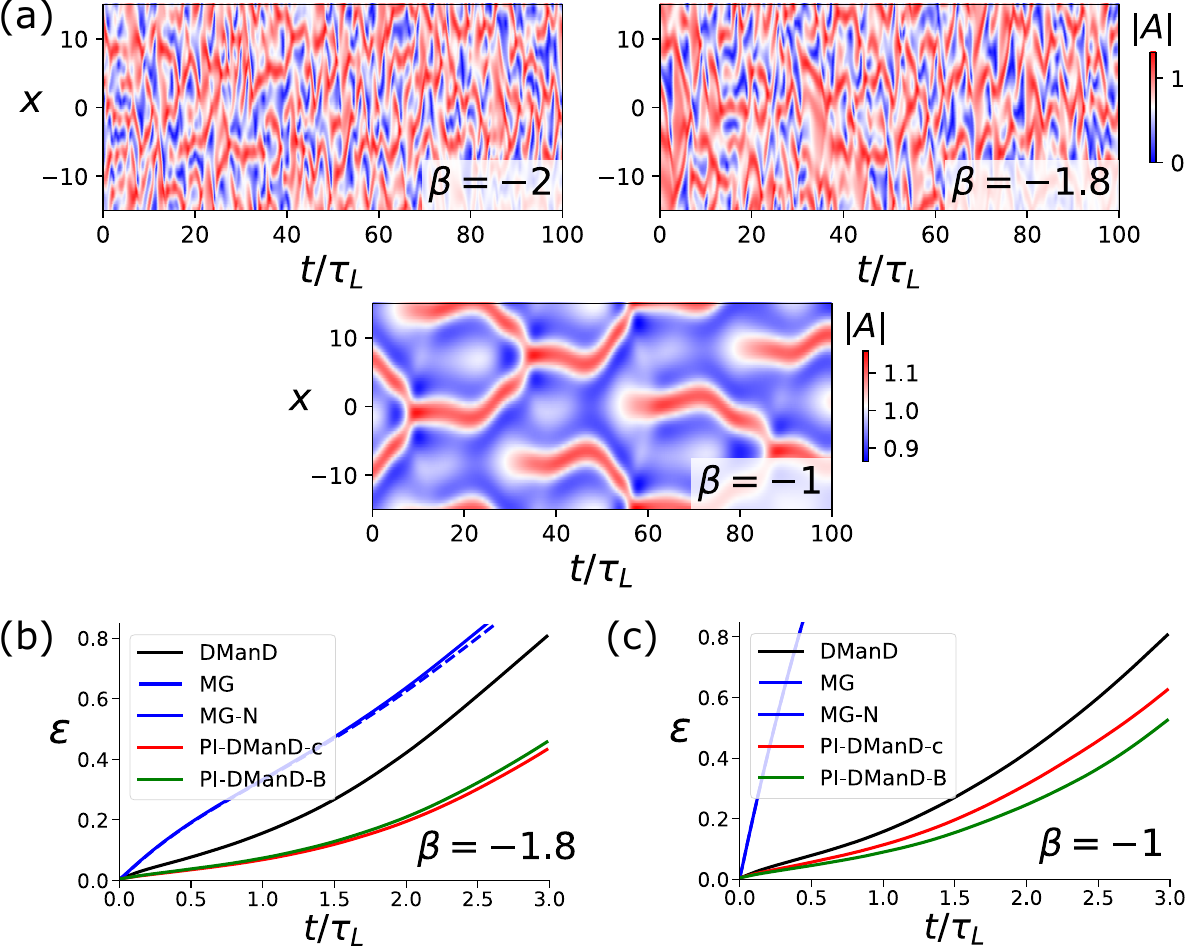}
	\caption{CGLE, $N=400,000$, $\tau=0.05$. Effects of using an erroneous parameter ($\beta=-1.8$ or $\beta=-1$) in the FOM to model data generated with $\beta=-2$. (a) Ground truth trajectory (top-left) and example trajectories for the two erroneous $\beta$ values (top-right and bottom); note the different colorbar axis limits. Ensemble-averaged short-time tracking error when using $\beta=-1.8$ (b) or $\beta=-1$ (c) in the FOM.}
	\label{fig:cgle_vary_beta}
\end{figure*}

We now test how well the hybrid methods can perform despite being informed with a FOM containing erroneous parameters. The results here parallel the KSE results in Section \ref{sec:kse_err_FOM}. We create a hybrid ROM for evolving the dataset generated with the correct parameters $\alpha=2$ and $\beta=-2$ when provided a FOM $f_\FOM$ with $\beta=-1.8$ or $\beta=-1$ (holding $\alpha=2$). For both $\beta=-2$ and $\beta=-1.8$, the dynamics lie in the defect-mediated turbulence regime \cite{Shraiman1992}; however, for $\beta=-1$, the dynamics instead lie in the phase-turbulence regime \cite{Shraiman1992}; example trajectories for these three values of $\beta$ are shown in Figure \ref{fig:cgle_vary_beta}(a). It bears repeating that the two erroneous $\beta$ values are only used in the $f_\FOM(u;\mu_\FOM)$ term appearing in Equation \ref{eq:MG} for MG (which also affects $g_{mn}$ learned by MG-N) --- no dataset is generated with the erroneous $\beta$ values.

We first compare short-time tracking performance. Figures \ref{fig:cgle_vary_beta}(b) and \ref{fig:cgle_vary_beta}(c) show the tracking error of all methods for $\beta=-1.8$ and $\beta=-1$, respectively. Unsurprisingly, the value of $\beta$ used in the FOM does not affect the predictive capability of DManD, since it does not use any prior knowledge of the physics. As was the case with KSE $L=20$ in Figure 4, MG and MG-N here also poorly track the true solution due to being informed with erroneous FOM parameters. (Note, the curves for MG and MG-N in Figure \ref{fig:cgle_vary_beta}(c) level off to around 1.5, and the vertical axis limits are shrunk to provide clarity.) As the accuracy of the prior decreases ($\beta$ strays further away from $-2$), the tracking error for PI-DManD-c and PI-DManD-B grow slightly but still remain lower than the error for DManD; however, the tracking error for PI-DManD-B is noticeably lower than that of PI-DManD-c for $\beta=-1$, which is in line with the KSE results in Section \ref{sec:kse_err_FOM} where only PI-DManD-B maintained a lower tracking error than DManD. Importantly, across both spatiotemporally chaotic datasets, the results show that including knowledge of the physics (even if inaccurate) helps in creating hybrid ROMs (PI-DManD-B especially) that perform better than solely data-driven ROMs (DManD) in short-time predictions.

\begin{figure*}
	\centering
	\includegraphics[width=1\linewidth]{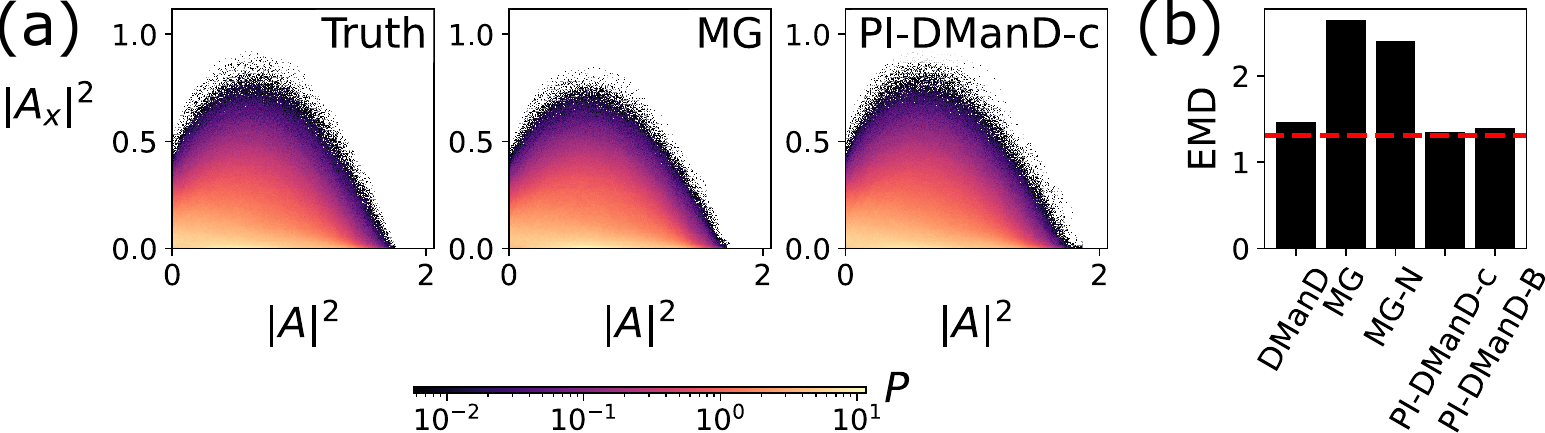}
	\caption{
    CGLE, N=400,000, $\tau=0.05$. Effects of using an erroneous parameter ($\beta=-1.8$) in the FOM to model the data generated with $\beta=-2$. (a) Comparison of joint pdfs for $\beta=-1.8$. (b) Earth movers distance (EMD) quantifying the difference between pdfs of the ground truth and model predictions for $\beta=-1.8$. Dashed red line is the baseline EMD calculated with the pdf of another ground truth trajectory. Joint pdfs for models not shown look similar to joint pdfs of models with similar EMD values.
    }
	\label{fig:cgle_vary_beta_lts}
\end{figure*}

Lastly, we investigate the long-time statistics of the model predictions by calculating the joint pdf of quantities $|A|^2$ and $|A_x|^2$. We plot the joint pdfs for the ground truth trajectory and the MG and PI-DManD-c model predictions in Figure \ref{fig:cgle_vary_beta_lts}(a) for $\beta=-1.8$.
The EMD values (see Equation \ref{eq:emd}) between pdfs of a model prediction and the ground truth test trajectory are provided in Figure \ref{fig:cgle_vary_beta_lts}(b).
(Again, the baseline EMD represented by the dashed red line in a bar plot is calculated between two ground truth trajectories, and pdfs for models not shown are similar to shown pdfs if they have similar EMD values.)
For $\beta=-1$, we report (but not show to reduce redundancy) that the EMD values for MG and MG-N were larger than 40 while the EMD values for the other three methods were as close to the baseline EMD of around 1.3 as for the $\beta=-1.8$ case.
When given an erroneous parameter value for the FOM, the two physics-informed MG and MG-N methods evolve the $\beta=-2$ dataset with a completely incorrect vector field, leading to poor reconstruction of long-time statistics --- this observation is more pronounced for the case of $\beta=-1$. As for the two hybrid methods, they capture the long-time statistics of these quantities about as well as DManD.

\end{document}